%% file: main.tex
\documentclass{article} 
\usepackage{iclr2022conference,times}

\input{math_commands.tex}

\usepackage{float}
\usepackage[utf8]{inputenc} 
\usepackage[T1]{fontenc}    
\usepackage{hyperref}       
\usepackage{url}            
\usepackage{booktabs}       
\usepackage{amssymb}
\usepackage{amsthm}
\usepackage{nicefrac}       
\usepackage{microtype}      
\usepackage{graphicx}
\usepackage{wrapfig}
\usepackage{subfigure}
\usepackage{booktabs}
\usepackage{algorithm,algcompatible}
\usepackage{enumitem}
\usepackage{multirow}
\usepackage{threeparttable}
\usepackage{textcomp}

\makeatletter
\DeclareRobustCommand\onedot{\futurelet\@let@token\@onedot}
\DeclareRobustCommand\onedot{\futurelet\@let@token\@onedot}
\def\@onedot{\ifx\@let@token.\else.\null\fi\xspace}

\title{Integrating Large Circular Kernels into CNNs through Neural Architecture Search} 

\author{Kun He\thanks{The first three authors contribute equally.}, Chao Li\thanks{Corresponding author.}, Yixiao Yang \\
School of Computer Science and Technology,\\
  Huazhong University of Science and Technology\\ 
  Wuhan, 430074, China \\ 
  \texttt{\{brooklet60, d201880880, m201973180\}@hust.edu.cn} \\
 \And
   Gao Huang \\
  Department of Automation,\\
  Tsinghua University\\
  Beijing, 100084, China \\
   \texttt{gaohuang@tsinghua.edu.cn} \\ 
 \And
   John E. Hopcroft \\
   Computer Science Department, \\
   Cornell University\\
   Ithaca, USA \\
   \texttt{jeh@cs.cornell.edu}  
}

\iclrfinalcopy 
\begin{document}
\maketitle
\begin{abstract}
The square kernel is a standard unit for contemporary CNNs, as it fits well on the tensor computation for convolution operation. However, the retinal ganglion cells in the biological visual system have approximately concentric receptive fields. Motivated by this observation, we propose to use circular kernel with a concentric and isotropic receptive field as an option for the convolution operation. We first propose a simple yet efficient implementation of the convolution using circular kernels, and empirically show the significant advantages of large circular kernels over the counterpart square kernels. We then expand the operation space of several typical Neural Architecture Search (NAS) methods with the convolutions of large circular kernels. The searched new neural architectures do contain large circular kernels and outperform the original searched models considerably. Our additional analysis also reveals that large circular kernels could help the model to be more robust to the rotated or sheared images due to their better rotation invariance. Our work shows the potential of designing new convolutional kernels for CNNs, bringing up the prospect of expanding the search space of NAS with new variants of convolutions.

\end{abstract}

\section{Introduction}
\label{sec:intro}
\input{01Intro}

\section{Related Works}
\label{sec:related_work}
\input{02RW}

\section{Circular Kernels for Convolution}
\label{sec:method}
\input{03Method}
\section{Experiments}
\label{sec:experiments}
\input{04ExpNew}

\section{Further Analysis}
\label{sec:analysis}
\input{05Analysis}

\section{Conclusion}
\label{sec.Conclusion}
In this work, we propose a new concept of circular kernel that could be an alternative option for the convolution operation in contemporary CNNs. The circular kernel exhibits approximately isotropic property and better rotation invariance because of the concentric and isotropic receptive field. 
We propose an simple yet efficient implementation of the convolution of circular kernels and reveal that the model with circular kernels has an optimization path different from that of the counterpart model with square kernels. 
Based on the increasing advantages of circular kernels over the counterpart square kernels 
with the increment of kernel size, 
we expand the operation space in several representative NAS methods with convolutions of large circular kernels because NAS enables large circular kernels to locate in proper position. We show that the searched architecture contains large circular kernels and outperforms the original architecture containing merely square kernels, and report state-of-the-art classification accuracy on benchmark datasets. 
Our work shows the potential of designing new shapes of convolutional kernels for CNNs, and bringing up the prospect of expanding the search space of NAS using variant of kernels that perform averagely as a whole in manual architecture but have extraordinary performance in the proper position in NAS. 

\clearpage
\bibliographystyle{splncs04}
\bibliography{egbib}

\clearpage
\appendix
\section{Appendix}
\label{sec.Appendix}
\input{06Appendix}

%
%

\end{document}

%% file: math_commands.tex
\def\eg{\emph{e.g}\onedot}

\def\etc{\emph{etc}\onedot} 
 
\def\etal{\emph{et al}.}

\makeatother

\usepackage{xspace}

\usepackage{color,xcolor}

\newcommand{\BLUE}[1]{\textcolor{blue}{#1}}
\newcommand{\red}[1]{\textit{\textcolor{red}{#1}}}

\newcommand{\real}{\mathbb{R}}

\usepackage{amsmath,amsfonts,bm}

\def\eqref#1{equation~\ref{#1}}
\def\Eqref#1{Equation~\ref{#1}}

\def\1{\bm{1}}

\def\vd{{\bm{d}}}

\def\vj{{\bm{j}}}

\def\vr{{\bm{r}}}

\def\mB{{\bm{B}}}

\def\mI{{\bm{I}}}

\def\mO{{\bm{O}}}

\def\mW{{\bm{W}}}

\DeclareMathAlphabet{\mathsfit}{\encodingdefault}{\sfdefault}{m}{sl}
\SetMathAlphabet{\mathsfit}{bold}{\encodingdefault}{\sfdefault}{bx}{n}

\def\sD{{\mathbb{D}}}

\def\sR{{\mathbb{R}}}
\def\sS{{\mathbb{S}}}



%% file: 01Intro.tex
The square convolution kernel has been regarded as the standard and core unit of Convolutional Neural Networks (CNNs) since the first recognized CNN of \textit{LeNet}  proposed in 1989~\citep{lecun1998gradient}, and especially after \textit{AlexNet}~\citep{krizhevsky2012imagenet} won the ILSVRC (ImageNet Large Scale Visual Recognition Competition) in 2012. Since then, various variants of convolution kernels have been proposed, including  separable convolution~\citep{chollet2017xception}, dilated convolution~\citep{yu2015multi}, deformable convolution~\citep{jeon2017active,dai2017deformable,zhu2019deformable,gao2019deformable}, \etc. 
Inspired by the fact that the retinal ganglion cells in the biological visual system have approximately concentric receptive fields (RFs)~\citep{hubel1962receptive,simoncelli2001natural,mutch2008object}, we propose the concept of circular kernels for the convolution operation\footnote{Code: https://github.com/JHL-HUST/CircularKernel.}. As shown in Fig. \ref{fig:3-5-7}, a $K \times K$ circular kernel is defined as a kernel that evenly samples $K^2$ pixels on the concentric circles to form a circular receptive field. 

Besides the similarity to biological RFs, we observe that the circular kernel provides many advantages over the square kernel. 
First, the receptive field of a kernel is traditionally expected to be isotropic to fit thousands of uncertain symmetric orientations of the input feature maps, either globally or locally. An \textit{isotropic} kernel means the kernel samples evenly in different directions of the RFs. The circular kernel is roughly isotropic and rotation-invariant, whereas a square kernel is symmetric only in a few orientations. 
Second, Luo \etal~\citep{luo2017understanding} indicate that the effective RF of a square kernel has a Gaussian distribution which is in a nearly circular shape. It indicates that the meaningful weights are sparse at the four corners of large square kernels or stacked $3 \times 3$ square kernels. Compared to pruning these diluted parameters during the fine-tuning stage~\citep{han2015learning}, directly constructing kernels with the same shape of effective RFs is probably more effective.
\par
 
\input{Figure/3-5-7}
 
One cornerstone 
of the rationality of employing circular kernels is the isotropic property of circles. However, a $3 \times 3$ circular kernel is not really in circular shape as it only samples nine pixels with a similar arrangement to the square kernel. If we build the circular kernels in larger kernel size, as illustrated in Fig. \ref{fig:3-5-7}, we can see that a larger circular kernel has a more round receptive field and is more distinct from the corresponding square kernel. Our follow-up experiments also demonstrate that the circular kernels exhibit significant advantages over the square kernels on larger kernel sizes.

The $3 \times 3$ square kernels have become the mainstream of the CNN units since the work of VGG~\citep{simonyan2014very} suggests that a larger square kernel could be substituted by several $3 \times 3$ square kernels utilizing fewer parameters. In recent years, however, the functions of larger square kernels have been considered underestimated, as almost all the powerful models generated by Neural Architecture Search (NAS)~\citep{zoph2018learning,liu2018progressive,DBLP:conf/iclr/XuX0CQ0X20,nayman2019xnas} contain large square kernels, and many manually designed neural architectures also contain large square kernels~\citep{he2016deep,peng2017large,li2019selective}. The recent 
success of ConvNeXt~\citep{liu2022convnet} over Swin transformer~\citep{liu2021swin} 
also shows that 
increasing the kernel size can significantly improve the performance. Compared to small kernels, large kernels have received insufficient attention despite their vast range of applications. 
Hence, we introduce convolutions with large circular kernels 
as an option for the CNN units, especially for NAS. 
\input{Figure/main_fig}

The mainstream CNNs are manually developed and optimized on the $3 \times 3$ square kernels. 
So variants of kernels encounter significant resistance to outperform $3 \times 3$ square kernels on the existing popular architectures. NAS aims to design a neural architecture that performs best under limited computing resources in an automated manner~\citep{ren2020comprehensive}. It creates a level playing field for different types of operations in the search space. In manually designed architectures, the network typically contains the same unit for all layers (e.g., $3 \times 3$ square kernels) because 
it is hard to 
arrange them in different layers appropriately if we have several different units. For the convolution operation, although it is hard to substitute the standard convolution with the variants in all layers of the typical manual architectures, NAS enables the variants to exist in the proper position as a part of the overall network. Then some special variants are likely to outperform the standard operations if being located in the right position. Consequently, although existing NAS methods have achieved superior performance, 
their search space that only contains popular operations used in manual architectures seems conservative.


In this work, we propose to use the convolution of large circular kernels with a concentric and isotropic receptive field as an option for the search space of NAS methods.
As shown in Fig.~\ref{fig:cifar},
by simply substituting the $3\times3$ square kernels with $3\times3$ circular kernels in manual CNN architectures, the performance of the modified network after training could be on par with the original network, even though the original one is designed and hence optimized manually on square kernels.
Moreover, with the increment of kernel size on the modified models adapted on typical manual models, although the overall performance declines for both square kernels and circular kernels, the circular kernels actually exhibit significantly increasing advantages over the counterpart square kernels. 

Our preliminary experiments inspire us to expand the search space of NAS methods with convolutions of large circular kernels. In this way, NAS 
may enable the convolutions of large circular kernels to be located in \textbf{proper position} to outperform the standard operations. 
Note that many works have shown that simply enlarging the operation space could be detrimental to the final results~\citep{DBLP:conf/iclr/YuSJMS20,DBLP:journals/pami/ZhangHWXP21,ci2020evolving}. 
Hence, our method of expanding the operation space with convolutions of large circular kernels is beneficial for the NAS.

Our main contributions are as follows:
\begin{itemize}[leftmargin=20pt]
	\vspace{-1.5mm}
	\item We first propose to use the circular kernel with a concentric and isotropic receptive field as an option for the convolution operation, especially for the convolution of kernels with larger kernel size.
	\item We propose a simple yet efficient implementation for the convolution of the circular kernel, enabling it to work together seamlessly with any CNNs with 
	little extra time. We also show that the circular kernel has an optimization path different from that of the square kernel.
	\item We propose to use the convolution of large circular kernels as an option for the search space of NAS methods because they enable large circular kernels to be located in proper position, and our experiments show that the searched architectures contain large circular kernels and outperform the original ones containing only square kernels. 
	\item Our study reveals an important phenomenon that the variants, that perform averagely in manual architecture because of inherent mode of thinking, probably have extraordinary performance in neural architecture search. We emphasize that the search space can be expanded as new designs emerge.
\end{itemize}

%% file: Figure/3-5-7.tex
\begin{figure*}[tb]
    \centering
    \includegraphics[width=0.7\linewidth]{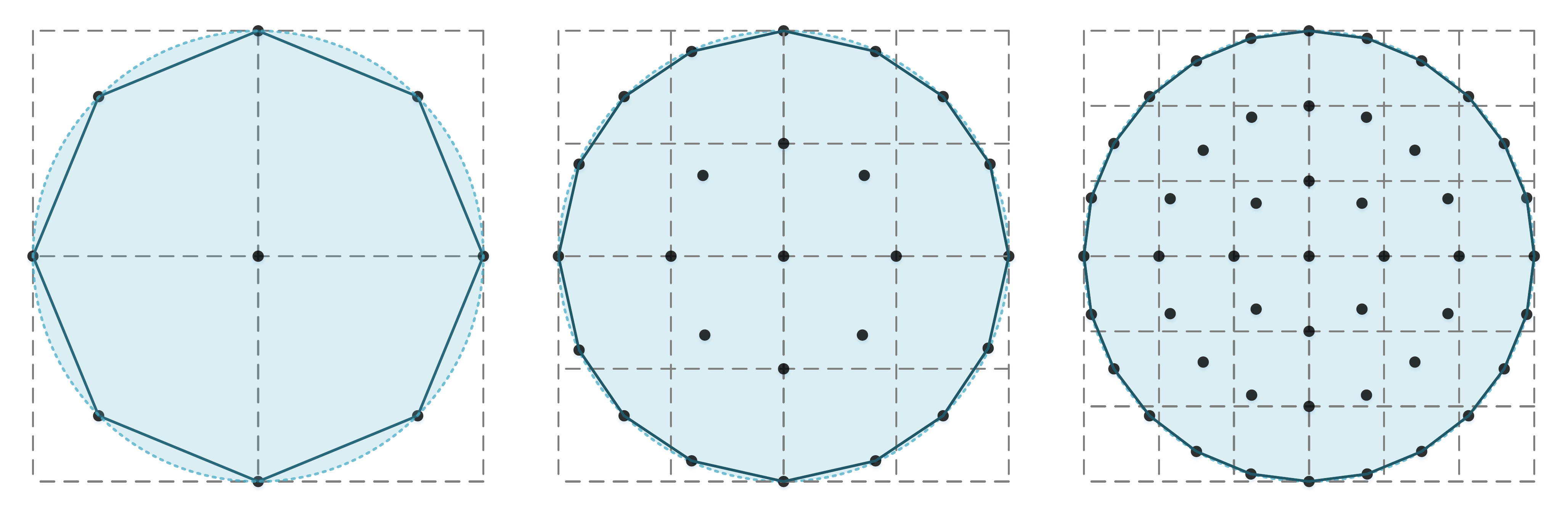}  
    \caption{The receptive fields and sampling points of circular kernels in size $k \in \{3,5,7\}$. The intersection of dashed lines are the sampling points of square kernels in size $k \in \{3,5,7\}$. All circular receptive fields are concentric and approximately isotropic. A larger circular kernel has a more round receptive field}
    \label{fig:3-5-7}
\end{figure*}

%% file: Figure/main_fig.tex
\begin{figure*}[tb]
\centering
\subfigure[\small{ResNet}]
{\label{fig:res}\includegraphics[width=0.242\linewidth]{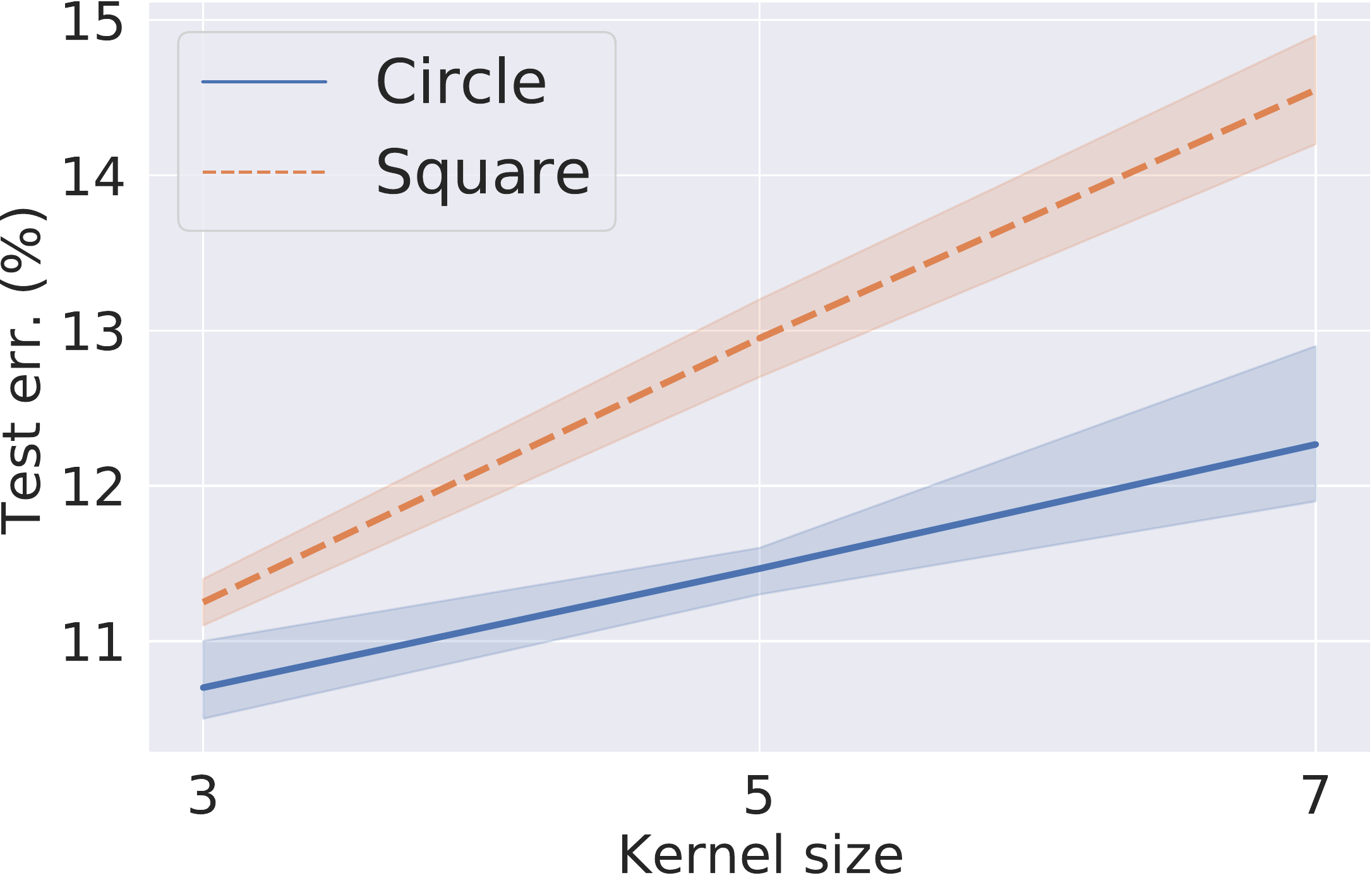}}
\subfigure[\small{VGG}]
{\label{fig:vgg}\includegraphics[width=0.242\linewidth]{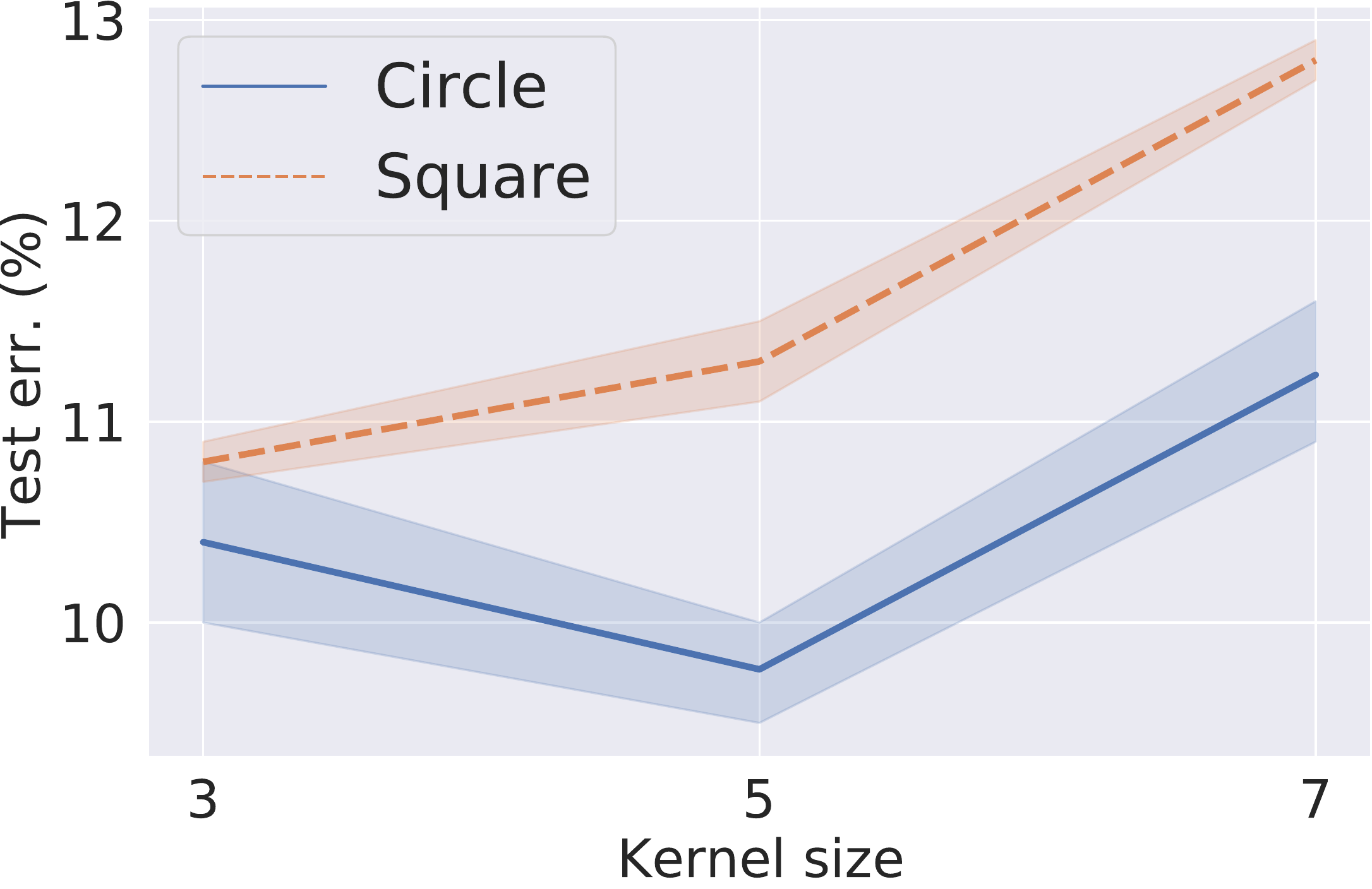}}
\subfigure[\small{WRNCifar}]
{\label{fig:wrn}\includegraphics[width=0.242\linewidth]{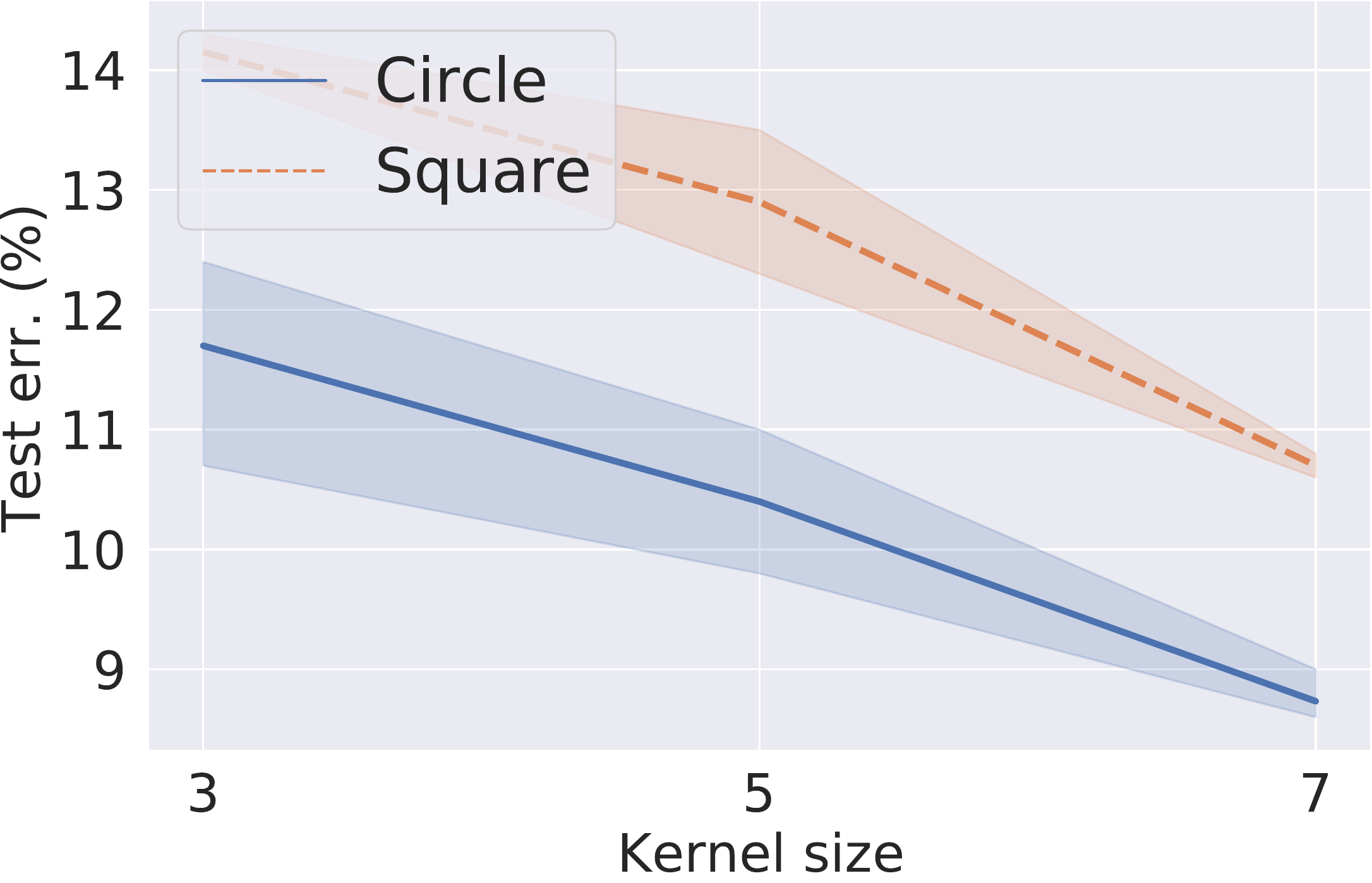}}
\subfigure[\small{DenseNetCifar}]
{\label{fig:dense}\includegraphics[width=0.242\linewidth]{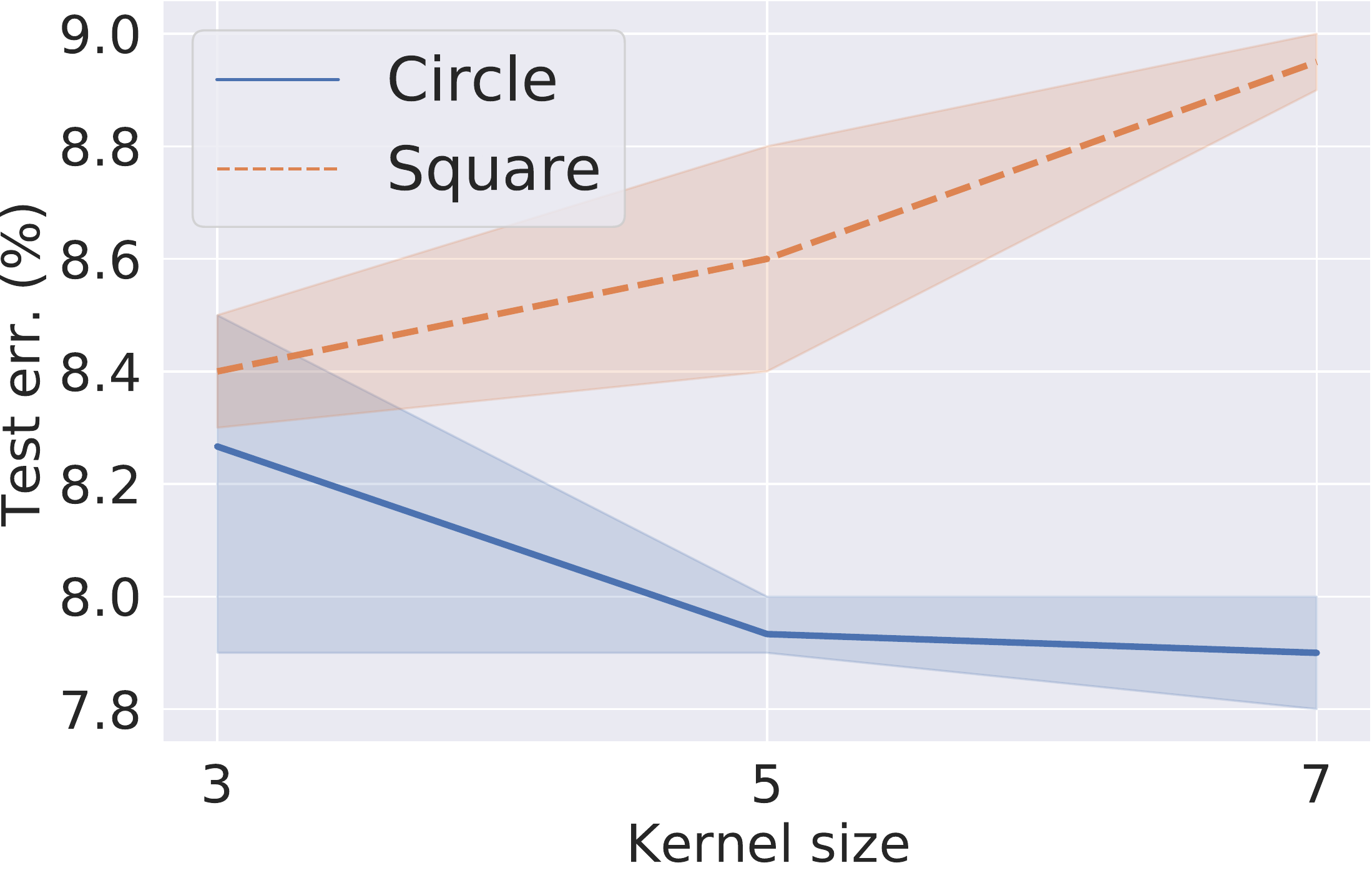}}\\
\subfigure[\small{ResNet}]{\label{fig:res100}\includegraphics[width=0.242\linewidth]{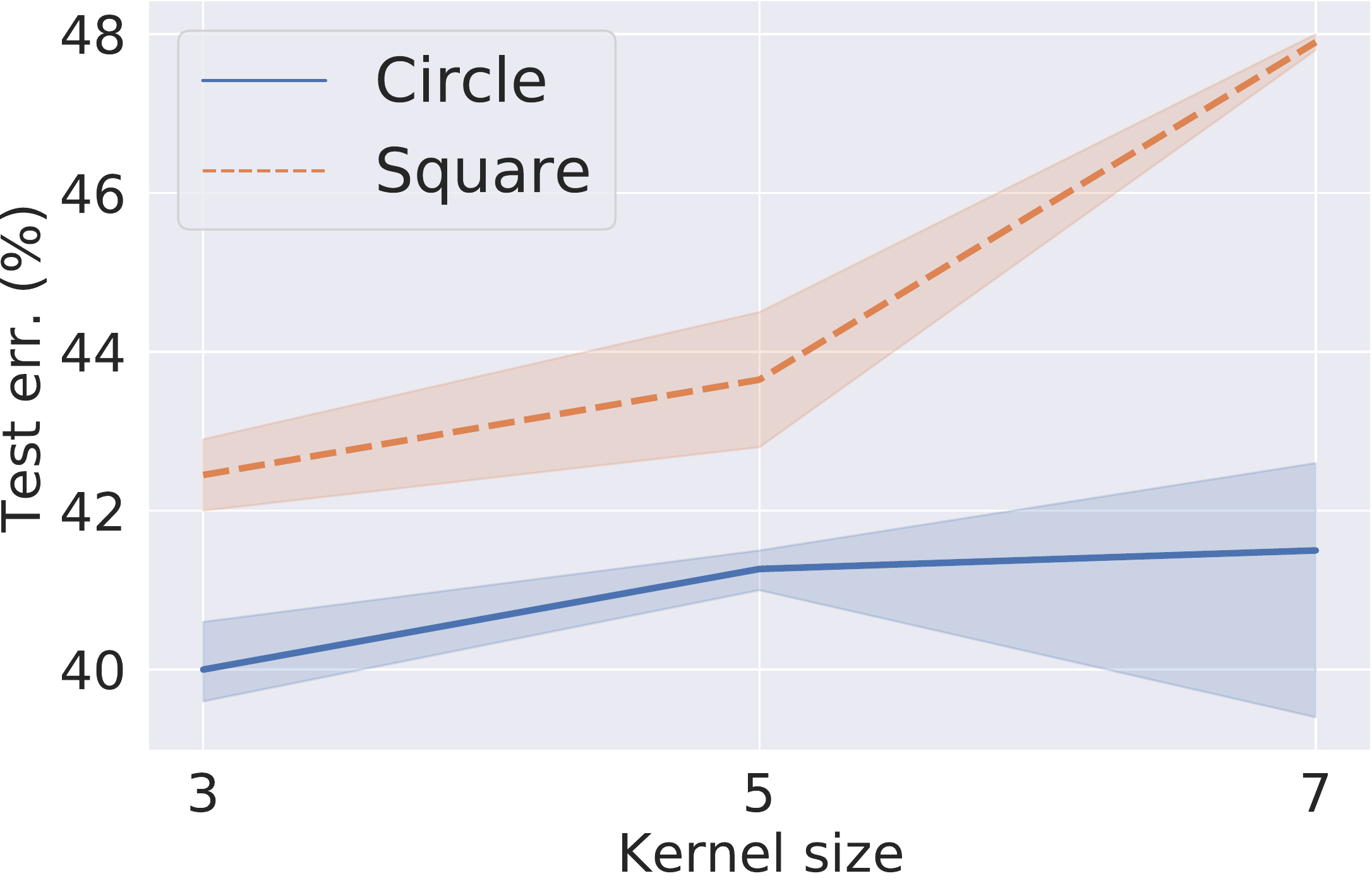}}
\subfigure[\small{VGG}]{\label{fig:vgg100}\includegraphics[width=0.242\linewidth]{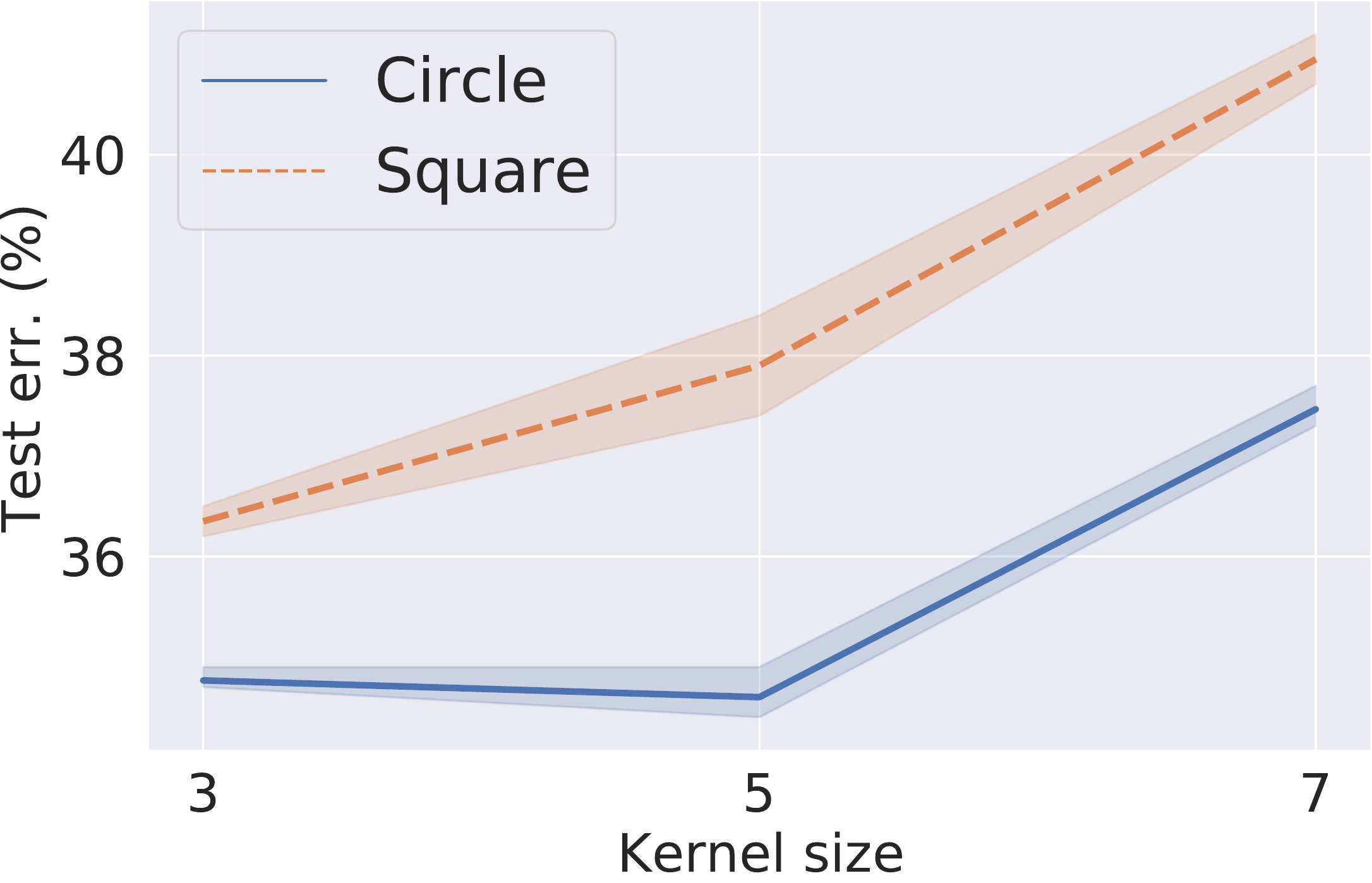}}
\subfigure[\small{WRNCifar}]{\label{fig:wrn100}\includegraphics[width=0.242\linewidth]{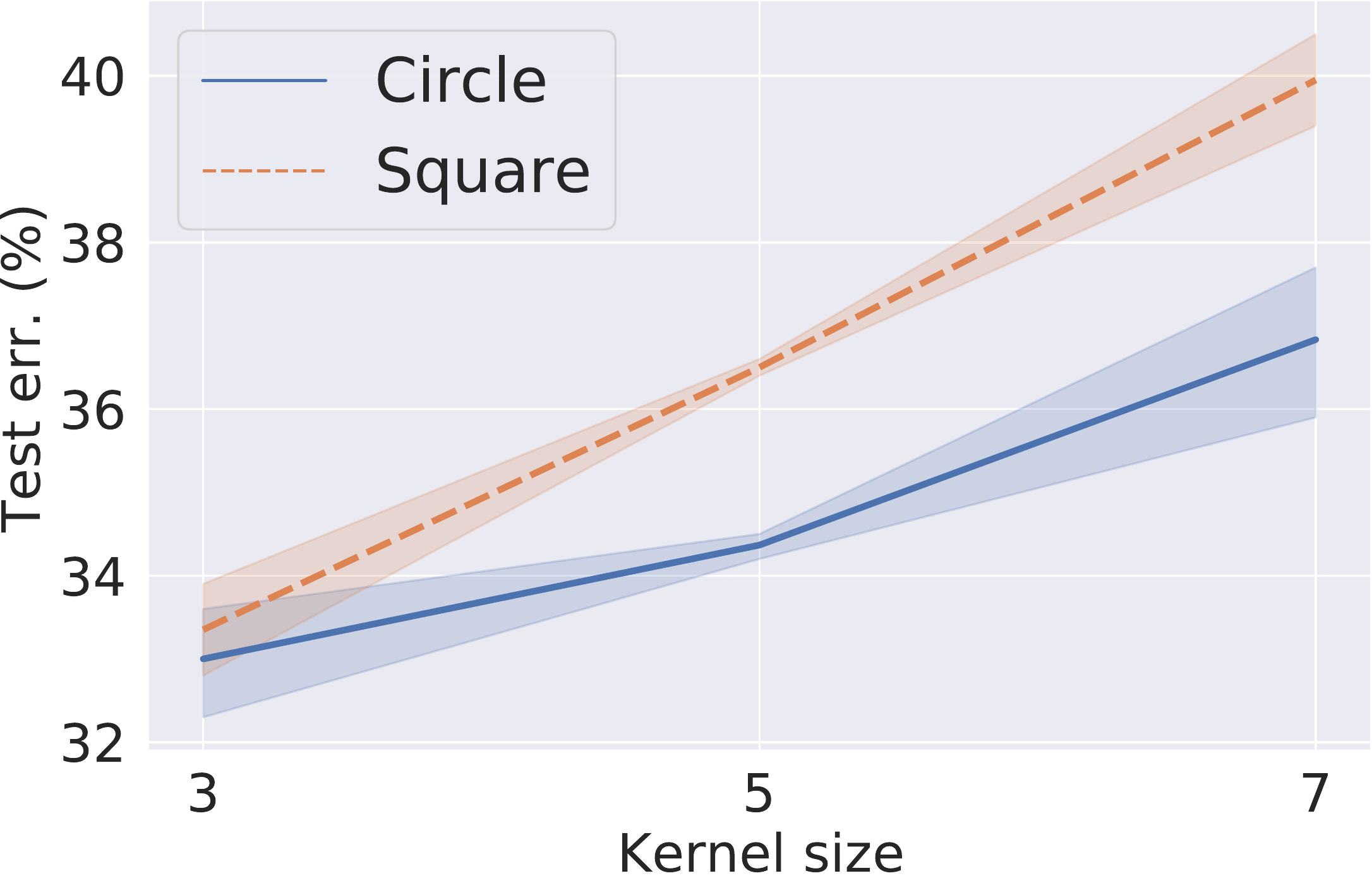}}
\subfigure[\small{DenseNetCifar}]{\label{fig:dense100}\includegraphics[width=0.242\linewidth]{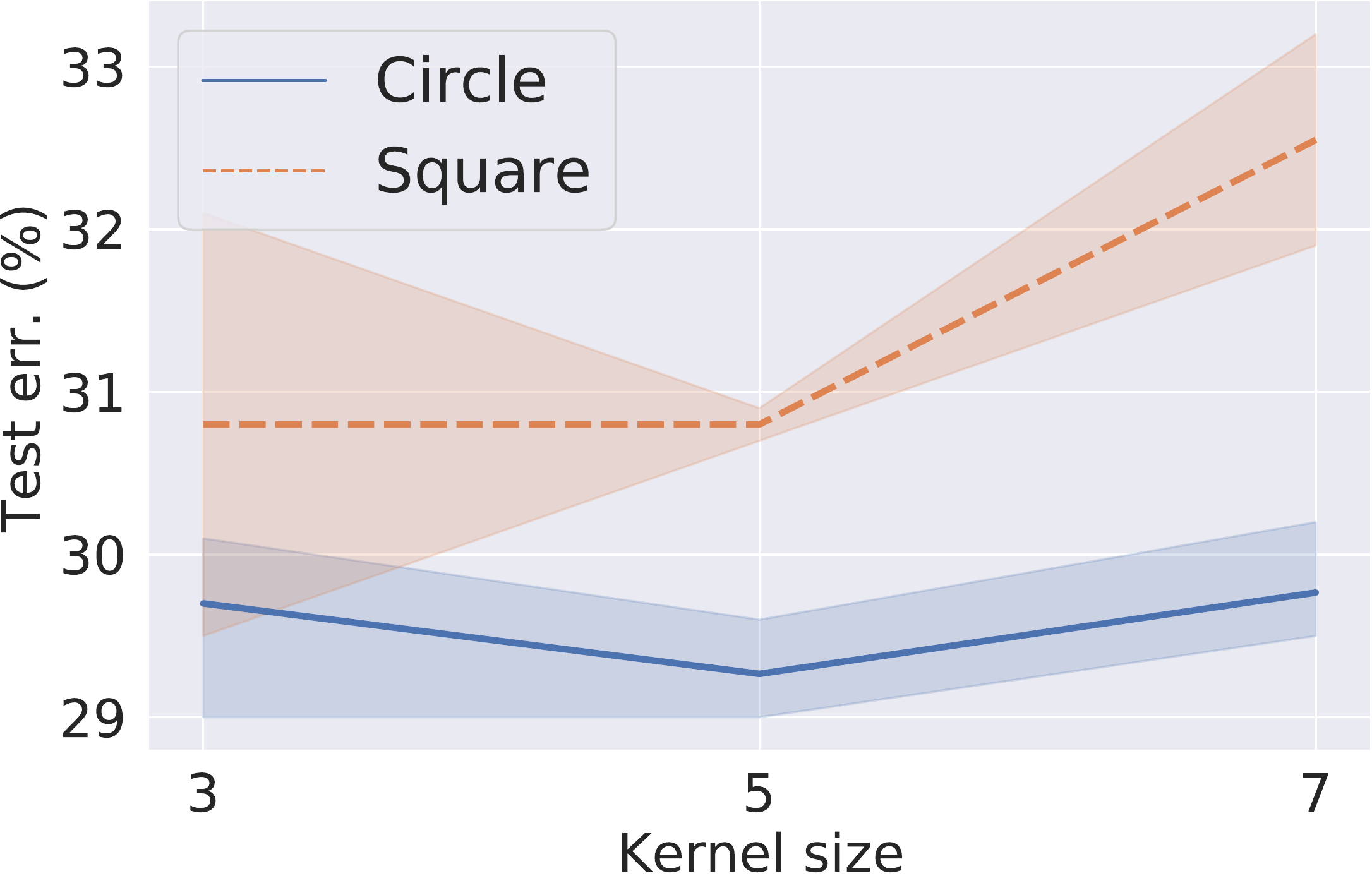}}

\caption{Test error ($\%$) of the baselines with square kernels (dashed lines) and the corresponding circular kernel (solid lines) versions in kernel size $k \in \{3,5,7 \}$ on CIFAR-10 (\textit{top}) and CIFAR-100 (\textit{bottom}). For a fair comparison, we use the original data without data augmentation (best viewed in color)}
\label{fig:cifar}
\end{figure*}


%% file: 02RW.tex
Understanding and exploring the convolution units has always been an essential topic in the field of deep learning. 
In this section, we review the previous primary efforts on the design of convolution kernel and CNN architecture, then show how our work differs.    
\vspace{1ex}

\textbf{Convolution Kernel Design.}
The grouped convolution uses a group of convolutions (multiple kernels per layer) to allow the network to train over multi-GPUs~\citep{krizhevsky2012imagenet}.
The depthwise separable convolution decomposes a standard convolution into a depthwise convolution followed by a pointwise convolution~\citep{chollet2017xception}. 
The spatially separable convolution decomposes a $K \times K$ square kernel into two separate units, a $K \times 1$ kernel and a $1 \times K$ kernel~\citep{mamalet2012simplifying}.
The dilated convolution is a type of convolution that “inflate” the kernel by inserting holes between the kernel elements~\citep{yu2015multi}. 
All the above variants consider large kernels but inherit the square kernel in general.
\vspace{1ex}

In contrast, the deformable convolution~\citep{dai2017deformable,zhu2019deformable} allows the shape of the receptive field to be learnable based on the input feature maps to provide flexibility, but it needs to take considerable extra parameters and computation overhead. Similarly, the deformable kernel~\citep{gao2019deformable} resamples the original kernel space while keeping the receptive field unchanged. There are also many interesting variants with special shapes, including quasi-hexagonal convolution~\citep{sun2016design}, blind-spot convolution~\citep{krull2019noise2void}, asymmetric convolution~\citep{ding2019acnet}, \etc. The above variants change the kernel shape but ignore large kernels.

In the early stage of CNN design, the kernel size gradually evolves from large to small. In AlexNet, large kernels (e.g., $11\times 11$, $5\times 5$) are used together with $3 \times 3$ kernels. Subsequently, VGG~\citep{simonyan2014very} suggests that a large kernel could be substituted by several $3 \times 3$ kernels utilizing fewer parameters. Then, the smallest $1 \times 1$ kernels are proposed for dimension reduction and efficient low dimensional embedding~\citep{szegedy2015going}. Recently, due to the emergence of NAS, large kernels (e.g., $5\times 5$, $7\times 7$) 
have been reemerged and attracted the researchers' attention, 
and become one of the standard units for the searched CNNs. ProxylessNAS~\citep{cai2018proxylessnas} argues that large kernels are beneficial for CNNs to preserve more information for the downsampling. 

\textbf{CNN Architecture Design.} 
Since AlexNet achieved fundamental progress in the ILSVRC-2012 image classification competition~\citep{krizhevsky2012imagenet}, a number of outstanding manual CNN structures emerged, including VGG~\citep{simonyan2014very}, Inception~\citep{szegedy2015going}, ResNet~\citep{he2016deep}, DenseNet~\citep{huang2017densely}, \etc. However, designing the neural architecture heavily relies on researchers’ prior knowledge, but existing prior knowledge and inherent mode of thinking  
are likely to limit the discovery of new neural architectures to a certain extent. As a result, neural architecture search (NAS) was developed to search good CNN structures automatically.

NAS-RL~\citep{DBLP:conf/iclr/ZophL17} and MetaQNN~\citep{DBLP:conf/iclr/BakerGNR17} using reinforcement learning (RL) are considered pioneers in the field of NAS. 
Subsequently, evolution-based algorithms use an evolving process towards better performance to search for novel neural architectures~\citep{xie2017genetic,real2017large,liu2017hierarchical,elsken2018efficient}. To address the issue of high computational demand and time cost in the search scenario, the one-shot method constructs a super-net~\citep{brock2017smash,bender2018understanding}, which is trained once in search and then deemed as a performance estimator. Some studies sample a single path~\citep{guo2019single,li2019random,you2020greedynas} in a chain-based search space~\citep{hu2020dsnas,cai2019once,mei2019atomnas,yu2020bignas} to train the super-net. Another line of gradient-based methods~\citep{DBLP:conf/iclr/LiuSY19,chen2019progressive,chu2020fair,DBLP:conf/iclr/XuX0CQ0X20,DBLP:conf/icml/ChenH20,yang2021towards,DBLP:conf/iclr/ChuW0LWY21} employs the gradient optimization method to perform differentiable joint optimization between the architecture parameters and the super-net weights in a cell-based space efficiently. Some gradient-based methods have reduced the search time significantly to about $0.1$ GPU-days~\citep{DBLP:conf/iclr/XuX0CQ0X20,yang2021towards}.

The search space of all the above works contains convolutions with large kernels that extensively exist in the final searched architectures. However, all the large kernels in these works directly inherit the square shape of the standard $3 \times 3$ kernel. Moreover, all the above works only employ operations that are popular in manual architectures to their operation space. Our work proposes to use convolutions of large circular kernels, which are more distinctive to the counterpart square kernels, to enrich the operation space for automatic neural architecture search. In turn, NAS may enable the convolutions of large circular kernels to be located in proper position to outperform the standard operations.
As far as we know, this work is the first that involves the unpopular variants in the search space of NAS.

%% file: 03Method.tex
\input{Figure/square-circle}
This section introduces the circular kernel that evenly samples pixels on concentric circles to form circular receptive fields. We adopt bilinear interpolation for the approximation and re-parameterize the weight matrix by the corresponding transformation matrix to replace the receptive field offsets, thus the training takes an approximately equivalent amount of calculation compared to the standard square convolution. 
We then provide preliminary analysis on 
the transformation matrix during training.
In the end, we show how to incorporate large circular kernels into the NAS methods.

\subsection{Circular Kernel versus Square Kernel} 
Without loss of generality, we take the $3 \times 3$ kernel as an example. 
The receptive field $\sS$ of a $3 \times 3$ standard square kernel with dilation 1, 
as shown in Fig. \ref{fig:square-circle} (a),
can be presented as: 
\begin{equation} 
\begin{aligned}
\sS=&\{(-1,\ \ 1 ),(0,\ \ 1 ),(1,\ \ 1 ),(-1,\ \ 0 ),(0,\ \ 0 ), \\
     &(1,\ \ 0 ),(-1, -1),(0,\  -1),(1,\  -1)\},
\end{aligned} 
\end{equation}
where $\sS$ denotes the set of offsets in the neighborhood considering the convolution conducted on the center
pixel. By convolving an input feature map $\mI \in \real^{H \times W}$ with a kernel $\mW \in \real^{K \times K}$ of stride 1, we have an output feature map $\mO \in \real^{H \times W}$,
whose value at each coordinate $\vj$ is:
\begin{equation}
    \label{eq.standard_conv}
	\mO_{j} = \sum_{s \in \sS} \mW_{s} \mI_{j + s}.
\end{equation}
So we have $\mO = \mathbf{W} \otimes \mI$ where $\otimes$ indicates a typical 2D convolution operation used in CNNs.

In contrast, the receptive field of a $3 \times 3$ circular kernel can be presented as: 
\begin{equation}
\begin{aligned}
\sR=&\{(-\tfrac{\sqrt{2}}{2},\tfrac{\sqrt{2}}{2}),(0,\ \ 1),(\tfrac{\sqrt{2}}{2}, \tfrac{\sqrt{2}}{2}),
               (-1,\  0 ),
               (0,\ 0 ),\\ 
               &(1,\ \ 0),
               (-\tfrac{\sqrt{2}}{2}, -\tfrac{\sqrt{2}}{2}), (0, \ -1),(\tfrac{\sqrt{2}}{2},-\tfrac{\sqrt{2}}{2})\}.
\end{aligned}
\end{equation}
As shown in Fig. \ref{fig:square-circle} (b), we resample the input $\mI$ with a group of offsets to each discrete kernel position $\bm{s}$, denoted as $\{\Delta\vr\}$, to form the circular receptive field. The corresponding convolution becomes:
\begin{equation}
    \mO_{j} =  \sum_{\bm{s}\in \sS} \mW_{\bm{s}} \mI_{\vj+\bm{s}+\Delta\vr}.
\label{eq.circle_conv}
\end{equation}
In other words, the value of each entry is the sum of the element-wise products of the kernel weights and the corresponding pixel values in the circular receptive field. 
As the sampling positions of a circular kernel contains fractional positions, we employ bilinear interpolation to approximate the corresponding sampling values inside the circular receptive field: 
\begin{equation}
\mI_{\vr}=\sum_{\bm{s} \in \sS} \mathcal{B}(\bm{s},\vr) \mI_{\bm{s}},
\label{eq.bilinear_interpolation_x}
\end{equation}
where $\vr$ denotes a grid or fractional location in the circular receptive field, $\bm{s}$ enumerates all 
the grid locations in the corresponding square receptive filed, and $\mathcal{B}(\cdot,\cdot)$ is a two dimensional bilinear interpolation kernel. $\mathcal{B}$ can be separated into two one-dimensional kernels as $\mathcal{B}(\bm{s},\vr)=g(\bm{s}_x,\vr_x)\cdot g(\bm{s}_y,\vr_y)$, where $g(a,b)=max(0,1-|a-b|)$. 
So $\mathcal{B}(\bm{s},\vr)$ 
is non-zero and in (0,1) 
only for the nearest four grids $\bm{s}$ in $\sS$ around fractional location $\vr$, and $\mathcal{B}(\bm{s},\vr) = 1$ only for the corresponding grid $\bm{s}$ in $\sS$ for grid location $\vr$.

\subsection{Re-parameterization of the Weights}
\label{Re-parameterization}
Implementing the convolution of a circular kernel that can operate efficiently is not trivial. Considering when building the convolution of a circular kernel, as the offsets of the sampling points in a circular receptive field relative to a square receptive field are fixed, we extract the transformation matrix $\mB$ of the whole receptive field by arranging $\mathcal{B}$ of one pixel $\vr$ in \Eqref{eq.bilinear_interpolation_x}. 

Let $\hat{\mI}_{RF(\vj)} \in \real^{K^2 \times 1}$ and $\hat{\mW} \in \real^{K^2 \times 1}$ represent the resized receptive field centered on the location $j$ and the kernel, respectively. The standard convolution can be defined as $\mO_{j}=\hat{\mW}^{\top}\hat{\mI}_{RF(\vj)}$. 
Then the convolution of circular kernel can be defined as:  
\begin{equation}
\mO_{j}=\hat{\mW}^{\top}\left(\mB\hat{\mI}_{RF(\vj)}\right) = \left(\hat{\mW}^{\top}\mB\right)\hat{\mI}_{RF(\vj)},
\label{eq.conv}
\end{equation}
where $\mB\in \real^{K^2 \times K^2}$ is a fixed sparse coefficient matrix. 
Correspondingly, let $\mI \in \real^{H \times W}$, $\mO \in \real^{H \times W}$ and $\mW \in \real^{K \times K}$ respectively represent the input feature map,  output feature map and the kernel, 
the convolution of a circular kernel could be briefly defined as:
\begin{equation}
\mO=\mW \otimes \left(\mB\star\mI\right)=\left(\mW \star \mB \right)\otimes\mI, 
\label{eq.trans}
\end{equation}
where $\mB\star\mI$ 
is to change the square receptive field to circular receptive field.

In this way, we could apply an operation on the kernel weights once to have $\mW \star \mB $ before the kernel scans the input feature map.
Consequently, we do not need to calculate the offsets for each convolution as deformable methods do when the kernel scans the input feature map step by step
~\citep{jeon2017active,dai2017deformable,zhu2019deformable,gao2019deformable}. 
While calculating the receptive field offsets for each convolution is very time-consuming, the computational cost of operations on kernels is negligible compared to the gradient descent optimization. 

\subsection{Analysis on Transformation Matrix}
\label{transformation}
This subsection briefly concludes the analysis of the actual effect of the transformation matrix. 
For a circular kernel, let $\Delta \mW = \mW^{t+1}-\mW^t$. The squared value of a change on the output $\Delta \mO = \mO^{t+1}-\mO^t$ can be calculated as 
$\| \Delta \mO \|^2 = \left( \mB\star \mI \right)^{\top} \otimes \Delta \mW^{\top} \Delta \mW \otimes \left( \mB\star \mI \right)$, which can be transferred to
$\| \Delta \mO \|^2 = \mI^{\top} \otimes \left( \mB^{\top}\star \Delta\mW^{\top} \Delta \mW \star \mB  \right) \otimes\mI$. 
In contrast, $\Delta \widetilde\mO$ of the traditional convolutional layers is determined by $\Delta \mW^{\top} \Delta \mW$ and $\mI$. 
Hence, we can conclude that the transformation matrix $\mB$ affects the 
optimization paths of gradient descent. 
For detailed analysis, see the Supplementary. 
We also empirically demonstrate this claim in Section~\ref{sec:3x3-Kernel}.


\subsection{NAS with Large Circular Kernels}
Theoretically, we could expand the operation space of any NAS method with large circular kernels. 
As the one-shot methods yield significant advantage in the time cost over the reinforcement learning or evolutionary-based NAS methods, and the typical one-shot methods of gradient-based ones~\citep{DBLP:conf/iclr/LiuSY19,chen2019progressive,DBLP:conf/iclr/XuX0CQ0X20,yang2021towards} enable us to discover more complex connecting patterns, we adopt them as the baselines for incorporating the large circular kernels.

The search space for typical gradient-based methods is made up of cell-based microstructure repeats. Each cell can be viewed as a directed acyclic graph with $N$ nodes and $E$ edges, where each node $x^i$ represents a latent representation (\eg, a feature map), and each edge is associated with operations $o(\cdot)$ (\eg, {\it identity connection}, {\it sep\_conv\_3x3}) in the operation space $\mathcal{O}$. Within a cell, the goal is to choose one operation from $\mathcal{O}$ to connect each pair of nodes. 

Let a pair of nodes be $\left(i,j\right)$, where ${1}\leqslant{i}<{j}\leqslant{N}$, the core idea of typical gradient-based methods is to formulate the information propagated from $i$ to $j$ as a weighted sum over $\left|\mathcal{O}\right|$ operations as the mixed operation: 
\begin{small}
\begin{align}
	\bar{o}^{(i,j)}\left(w,\mathbf{x}_i\right) = \sum_{o \in \mathcal{O}} \frac{\exp(\alpha_o^{(i,j)})}{\sum_{o' \in \mathcal{O}} \exp(\alpha_{o'}^{(i,j)})} o\left(w_o^{(i,j)},\mathbf{x}_i\right),
	\label{eq:soft}
\end{align}
\end{small}
where $\mathbf{x}_i$ is the output of the $i$-th node, $\alpha_o^{(i,j)}$ is a hyper-parameter for weighting operation $o\!\left(\mathbf{x}_i\right)$, and $w_o^{(i,j)}$ is the weight. The entire framework is then differentiable to both layer weights in operation $o(\cdot)$ and hyper-parameters $\alpha_o^{(i,j)}$ in an end-to-end fashion.  
After that, a discrete architecture can be obtained by replacing mixed operations with most likely operations at the end of the search. 

The operation space $\mathcal{O}$ of typical gradient-based methods is: $3 \times 3$ and $5 \times 5$ separable convolutions, $3 \times 3$ and $5 \times 5$ dilated separable convolutions, $3 \times 3$ max pooling, $3 \times 3$ average pooling, identity, and \textit{zero}, which only considers convolutions or pooling that are popular in manually designed CNNs. 
Considering variants of convolutions may outperform the popular convolutions when they are located in the proper position, we add $5 \times 5$ circular separable convolutions and $5 \times 5$ circular dilated separable convolutions to the operation space, which are constructed by replacing the square kernels of the separable or dilated convolution with circular kernels.

%% file: Figure/square-circle.tex
\begin{figure*}[tb]
    \centering
    \includegraphics[width=0.75\textwidth]{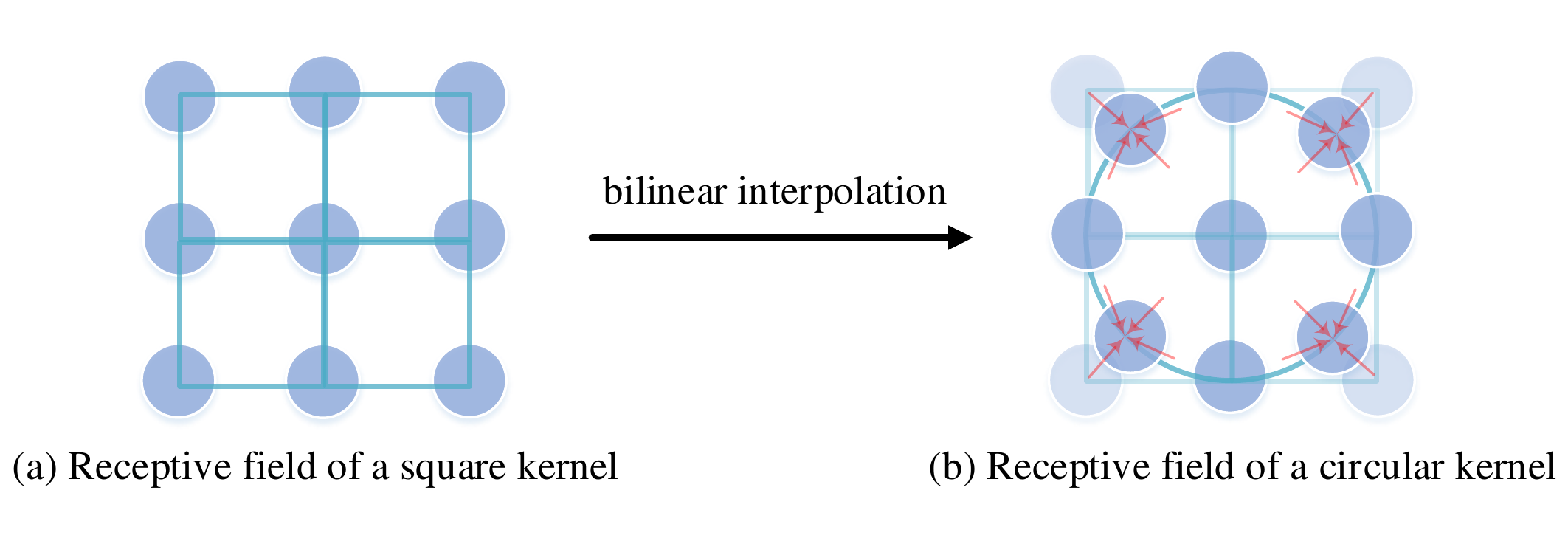}
    \vspace{-15pt}
    \caption{Approximation of a $3 \times 3$ circular kernel on a $3 \times 3$ square kernel 
    }
    \label{fig:square-circle}
\end{figure*}

%% file: 04ExpNew.tex

In this section, we empirically demonstrate the advantages of large circular kernels 
over large square kernels. 
Then we expand the search space of NAS with large circular kernels and apply the strategy described in several representative gradient-based NAS methods to search for new neural network architectures based on their ability of locating large circular kernels in proper position. Experimental results show that the searched architectures contain large circular kernels and outperform the original ones 
on both CIFAR-10 and ImageNet datasets. Detailed experimental setup can be found in the Supplementary. 

\subsection{The Advantages of Large Circular Kernels}
\label{sec:cifar}
We have shown in Fig.~\ref{fig:3-5-7} that a large circular kernel has a more round receptive field and is more distinguishable from the corresponding square kernel. We conjecture that the larger circular kernels would exhibit a more significant advantage over the square kernels if the circular kernels are helpful for deep learning tasks. To verify this hypothesis, we augment VGG~\citep{simonyan2014very}, ResNet~\citep{he2016deep}, WRNCifar~\citep{zagoruyko2016wide}, DenseNetCifar~\citep{huang2017densely}, and their circular kernel versions with larger kernel sizes and compare their performance on CIFAR-10 and CIFAR-100. For a fair comparison, we show the results of the original data without data augmentation. Results on CIFAR-10 and CIFAR-100 with standard data augmentation can be found in the Supplementary. We run the model 5 times for each dataset \& model setting to reduce the variance.  

As shown in Fig.~\ref{fig:cifar}, 
the performance of both the baselines and the corresponding circular kernel versions basically declines with the increment of the kernel size. Nevertheless, we see that the advantage of circular kernels over square kernels becomes more distinct for larger kernels. The average gain brought by circular kernels on four models is $1.1\%$ for kernel size of $3$, $2.0\%$ for kernel size of $5$, and $2.8\%$ for kernel size of $7$, indicating the superiority of large circular kernels. \par



\subsection{Searched Models with Large Circular Kernels}
\label{sec:NAS-search}
\input{Table/NAS-search-CIFAR10}

Although the models using large circular kernels surpass the counterpart models using large square kernels, the large kernels are not superior to standard $3 \times 3$ kernels as a whole in the manually designed models.
In this subsection, we incorporate the large circular kernels into the advanced gradient-based methods for neural architecture search so as to confirm the proper place of large circular kernels. 
DARTS~\citep{DBLP:conf/iclr/LiuSY19} is the first NAS method based on joint gradient optimization, 
and PC-DARTS~\citep{DBLP:conf/iclr/XuX0CQ0X20} is one of the best gradient-based methods. 
PC-DARTS enables a direct architecture search on ImageNet with only $3.8$ GPU-days while most other NAS methods can only search on CIFAR and then evaluate on ImageNet. 
Hence, we incorporate the convolutions with large circular kernels to the operation space of DARTS and PC-DARTS. Denote our newly searched architectures as DARTS-Circle and PC-DARTS-Circle, respectively. 

\input{Table/NAS-search-imagenet}

On CIFAR-10, the search and evaluation scenarios simply follow DARTS and PC-DARTS except for some necessary changes for a fair comparison. In the search scenario, the over-parameterized network is constructed by stacking $8$ cells ($6$ normal cells and $2$ reduction cells) for DARTS-Circle and PC-DARTS-Circle, and each cell consists of ${N}={6}$ nodes. In cell $k$, the first $2$ nodes are input nodes, which are the outputs of cells $k - 2$ and $k - 1$, respectively. Each cell's output is the concatenation of all the intermediary nodes. In the evaluation stage, the network comprises $20$ cells ($18$ normal cells and $2$ reduction cells), and each type of cell shares the same architecture. 

On ImageNet, following DARTS and PC-DARTS, the over-parameterized network starts with three convolution layers of stride 2 to reduce the input image resolution from $224 \times 224$ to $28 \times 28$. The cell architecture is the same with CIFAR-10. To reduce the search time, we randomly sample two subsets from the $1.3\mathrm{M}$ training set of ImageNet, with $10\%$ and $2.5\%$ images, respectively. 
In the evaluation stage, we apply the most popular \textit{mobile setting} where the input image size is fixed to be $224 \times 224$, and the number of multi-add operations does not exceed $600\mathrm{M}$~\citep{DBLP:conf/iclr/LiuSY19,DBLP:conf/iclr/XuX0CQ0X20,chen2019progressive}. 

The CIFAR-10 results for various convolutional architectures are presented in Table~\ref{NAS-search-CIFAR10}. Notably, both DARTS-Circle and PC-DARTS-Circle outperform the DARTS and PC-DARTS baselines with similar FLOPs, respectively. With AutoAugment~\citep{cubuk2018autoaugment}, PC-DARTS has a surprising performance of $2.15\%$ error rate. However, PC-DARTS-Circle can still boost the performance of PC-DARTS by $+0.13\%$ and achieves a state-of-the-art performance of $2.02\%$ error rate with only $0.1$ GPU-days.

The comparisons on ImageNet 
are summarized in Table~\ref{NAS-search-imagenet} and Fig.~\ref{fig:acc_flops}, illustrating that NAS methods have a better trade-off than the strong baseline of manual architecture.
In NAS methods, DARTS-Circle achieves a top-1/5 error of $25.9\%$/$8.1\%$, considerably outperforming the results of $26.7\%$/$8.7\%$ reported by DARTS with similar FLOPs. Limited to the mobile setting, we reduce the $14$ stacked cells used in PC-DARTS-Circle to $13$ stacked cells. 
\input{Figure/acc_flops}
However, PC-DARTS-Circle can still achieve a state-of-the-art top-1/5 error of $24.0\%$/$7.1\%$, slightly outperforming the results of $24.2\%$/$7.3\%$ reported by PC-DARTS. 
When PC-DARTS-Circle uses the same hyper-parameter of 14 stacked cells with PC-DARTS, denoted by PC-DARTS-Circle-v2, the top-1/5 error further reduces to $23.7\%$/$7.0\%$, which is the best result of differentiable architecture search approaches under DARTS-based search space as far as we know. The main difference between PC-DARTS-Circle and PC-DARTS is that the former contains large circular kernels. So we can conclude that large circular kernels are excellent candidates for NAS.
It is also worth noting that PC-DARTS-Circle outperforms NSENet, whose operation space contains $27$ traditional operations without convolutions of circular kernels and is much more than the $9$ operations employed in PC-DARTS-Circle. 
\par 

\input{Figure/searched_cells_PC_DARTS}

We visualize the searched normal cells and reduction cells of PC-DARTS-Circle on CIFAR-10 (\textit{left}) and ImageNet (\textit{right}) in Fig.~\ref{fig:searched_cells_PC_DARTS}. All other searched cells are shown in the Supplementary. 
Although large circular kernels only exist in a few layers
and mainly exist in the reduction cells, they have a significant impact on the overall network because the receptive field is stacked as the layers go deeper.

%% file: Table/NAS-search-CIFAR10.tex
\begin{table*}[t]
\centering
\caption{Comparison with state-of-the-art searched network architectures on CIFAR-10. $\ddagger$ denotes
model trained with AutoAugment~\cite{cubuk2018autoaugment}}
\small
\begin{threeparttable}[b]
\resizebox{0.8\textwidth}{!}{

\begin{tabular}{@{}lcccccc@{}}
\toprule
\multirow{2}{*}{\textbf{Architecture}} & \textbf{Test Err.} & \textbf{Params} & \textbf{Search Cost} & \multirow{2}{*}{\textbf{Search Method}} \\
&                            \textbf{(\%)} & \textbf{(M)} & \textbf{(GPU-days)} &\\
\midrule
DenseNet-BC~\cite{huang2017densely}                       & 3.46  & 25.6 & -    & manual \\
\midrule
NASNet-A + cutout~\cite{zoph2018learning}                 & 2.65  & 3.3  & 1800 & RL      \\
AmoebaNet-A + cutout~\cite{real2019regularized}          & 3.34$\pm$0.06 &  3.2  & 3150 & evolution \\
AmoebaNet-B + cutout~\cite{real2019regularized}           & 2.55$\pm$0.05 & 2.8  & 3150 & evolution \\
 
\midrule
P-DARTS + cutout~\cite{chen2019progressive}      & 2.50 &  3.4  & 0.3 & gradient-based \\
R-DARTS(L2) + cutout~\cite{bender2018understanding} & 2.95$\pm$0.21 & - & 1.6 & gradient-based \\
S-DARTS-ADV + cutout~\cite{DBLP:conf/icml/ChenH20} & 2.61$\pm$0.02 & 3.3 & 1.3 & gradient-based \\
Fair DARTS + cutout~\cite{chu2020fair} & 2.54$\pm$0.05 & 3.3 & - & gradient-based \\
EnTranNAS-DST + cutout~\cite{yang2021towards}  &2.48$\pm$0.08&3.2&0.1& gradient-based\\
DARTS$-$ + cutout~\cite{DBLP:conf/iclr/ChuW0LWY21} & 2.59$\pm$0.08 & 3.5 & 0.4 & gradient-based\\
\midrule
DARTS + cutout~\cite{DBLP:conf/iclr/LiuSY19}          & 2.76$\pm$0.09 &  3.3 & 0.4 & gradient-based \\
PC-DARTS + cutout$^\ddagger$~\cite{DBLP:conf/iclr/XuX0CQ0X20} &2.15$\pm$0.04 & 3.6  & 0.1& gradient-based \\
\midrule
\midrule
DARTS-Circle + cutout    & 2.62$\pm$0.08 &  3.9 & 0.4 & gradient-based \\
PC-DARTS-Circle + cutout$^\ddagger$ & 2.02$\pm$0.05 & 3.5  & 0.1& gradient-based \\
\bottomrule
\end{tabular}
}
\end{threeparttable}

\label{NAS-search-CIFAR10}
\end{table*}

%% file: Table/NAS-search-imagenet.tex
\begin{table*}[tb]
\centering
\caption{Comparison with state-of-the-art searched architectures on ImageNet (\textbf{mobile setting}). $\ddagger$: These architectures are searched on ImageNet directly, others are searched on CIFAR-10 or CIFAR-100 and transferred to ImageNet}

\begin{threeparttable}[b]
\resizebox{0.9\textwidth}{!}{
\begin{tabular}{@{}lcccccc@{}}
\toprule
\multirow{2}{*}{\textbf{Architecture}} & \multicolumn{2}{c}{\textbf{Test Err. (\%)}} & \textbf{Params} & \textbf{FLOPs} & \textbf{Search Cost} & \multirow{2}{*}{\textbf{Search Method}} \\
\cmidrule(lr){2-3}
&                            \textbf{top-1} & \textbf{top-5} & \textbf{(M)} & \textbf{(M)} & \textbf{(GPU-days)} &\\
\midrule
ResNet50~\cite{he2016deep} & 24.7 &-& 25.6 &4100 & - & manual \\
Inception-v1~\cite{szegedy2015going}          & 30.2 & 10.1 & 6.6  & 1448 &-    & manual \\
MobileNet~\cite{howard2017mobilenets}         & 29.4 & 10.5 & 4.2   & 569  &-    & manual \\
ShuffleNet 2$\times$ (v2)~\cite{ma2018shufflenet}    & 25.1 & - & $\sim$5    & 591  &-    & manual \\
\midrule
NASNet-A~\cite{zoph2018learning}              & 26.0 & 8.4  & 5.3   & 564  &1800 & RL \\
AmoebaNet-C~\cite{real2019regularized}        & 24.3 & 7.6  & 6.4   & 570  &3150 & evolution \\
FairNAS-A~\cite{chu2019fairnas}               & 24.7 &  -   & 4.6    & 388  & 12    &evolution \\  
\midrule
P-DARTS~\cite{chen2019progressive}                 & 24.4 & 7.4  & 4.9   & 557  & 0.3  & gradient-based \\
S-DARTS-ADV~\cite{DBLP:conf/icml/ChenH20}&25.2 & 7.8 & -&- & -&gradient-based \\
Fair DARTS~\cite{chu2020fair}$^\ddagger$ &24.4 &7.4 & 4.3&440&3.0&gradient-based \\
NSENet~\cite{ci2020evolving}                         & 24.5 & -  & 4.6   & 330&-  & gradient-based \\ 
\midrule

DARTS~\cite{DBLP:conf/iclr/LiuSY19}      & 26.7 & 8.7  & 4.7   & 574  &0.4    & gradient-based \\
PC-DARTS (CIFAR-10)~\cite{DBLP:conf/iclr/XuX0CQ0X20} &25.1&7.8&5.3& 586 & 0.1 & gradient-based \\
PC-DARTS (ImageNet)$^\ddagger$~\cite{DBLP:conf/iclr/XuX0CQ0X20}  & 24.2 &7.3  & 5.3  & 597  & 3.2 & gradient-based \\ 
\midrule
\midrule
DARTS-Circle      & 25.9 & 8.1  & 5.3   & 583    & 0.4 & gradient-based \\ 
PC-DARTS-Circle (CIFAR-10) &24.9&7.7&5.0&571&0.1&gradient-based \\
PC-DARTS-Circle (ImageNet)$^\ddagger$ &  24.0  &  7.1  & 5.5  & 599 & 3.2 & gradient-based \\ 
PC-DARTS-Circle-v2 (ImageNet)$^\ddagger$ &  23.7  &  7.0  & 5.7  & 639 & 3.2 & gradient-based \\ 
\bottomrule			
\end{tabular}}
\end{threeparttable}

\label{NAS-search-imagenet}
\end{table*}

%% file: Figure/acc_flops.tex
\begin{wrapfigure}{tbR}{0.5\columnwidth}
\vskip -5pt
	\centering
	\includegraphics[width=0.5\columnwidth]{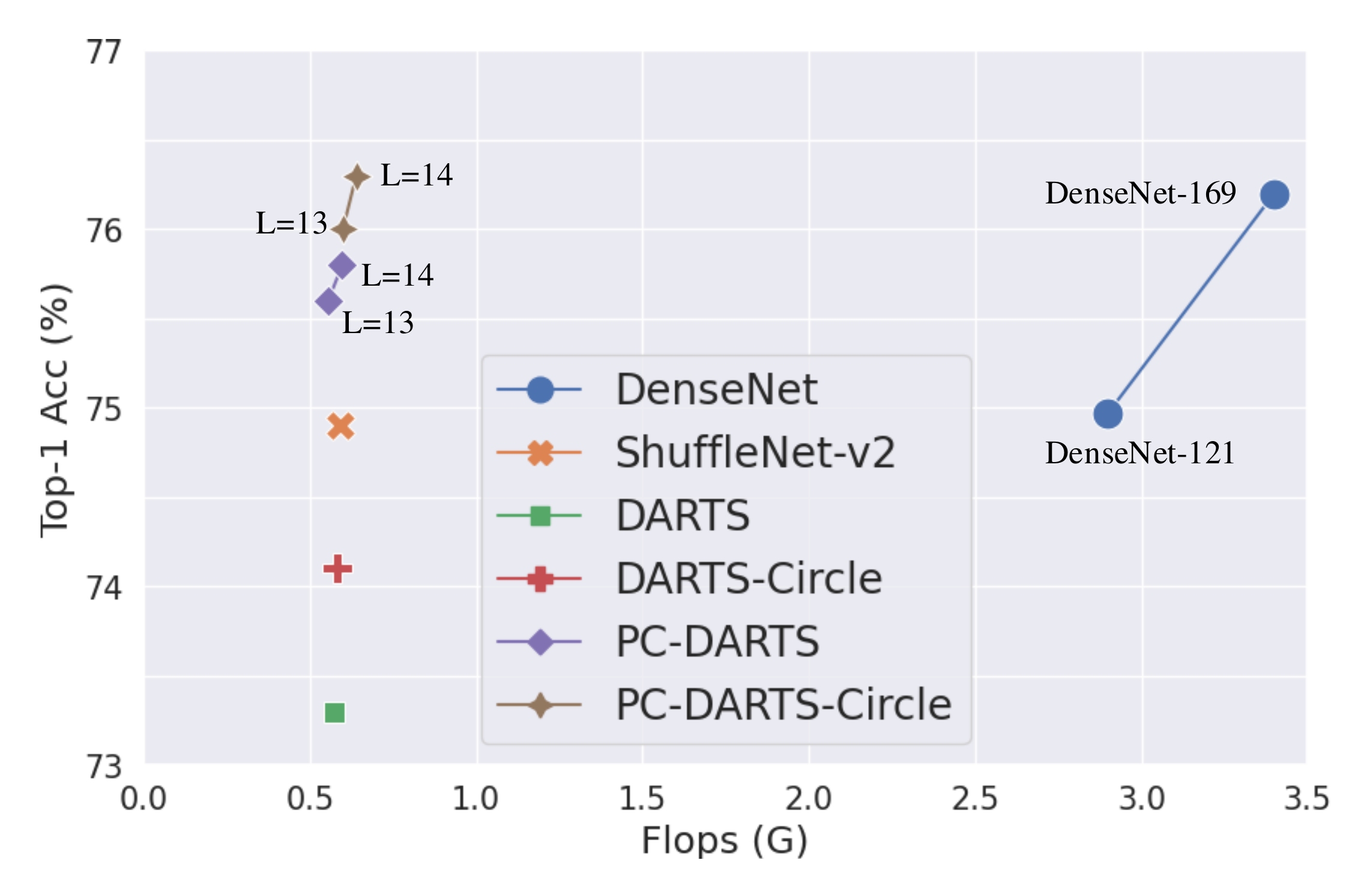}
	\caption{Comparison of top-1 accuracy on ImageNet with FLOPs (best viewed in color)}
	\label{fig:acc_flops}
	\vskip -5pt
\end{wrapfigure}

%% file: Figure/searched_cells_PC_DARTS.tex
\begin{figure*}[tb]
\centering
\subfigure[The normal cell on CIFAR-10]{\label{pc_im_ncells_s1}\includegraphics[width=0.42\linewidth]{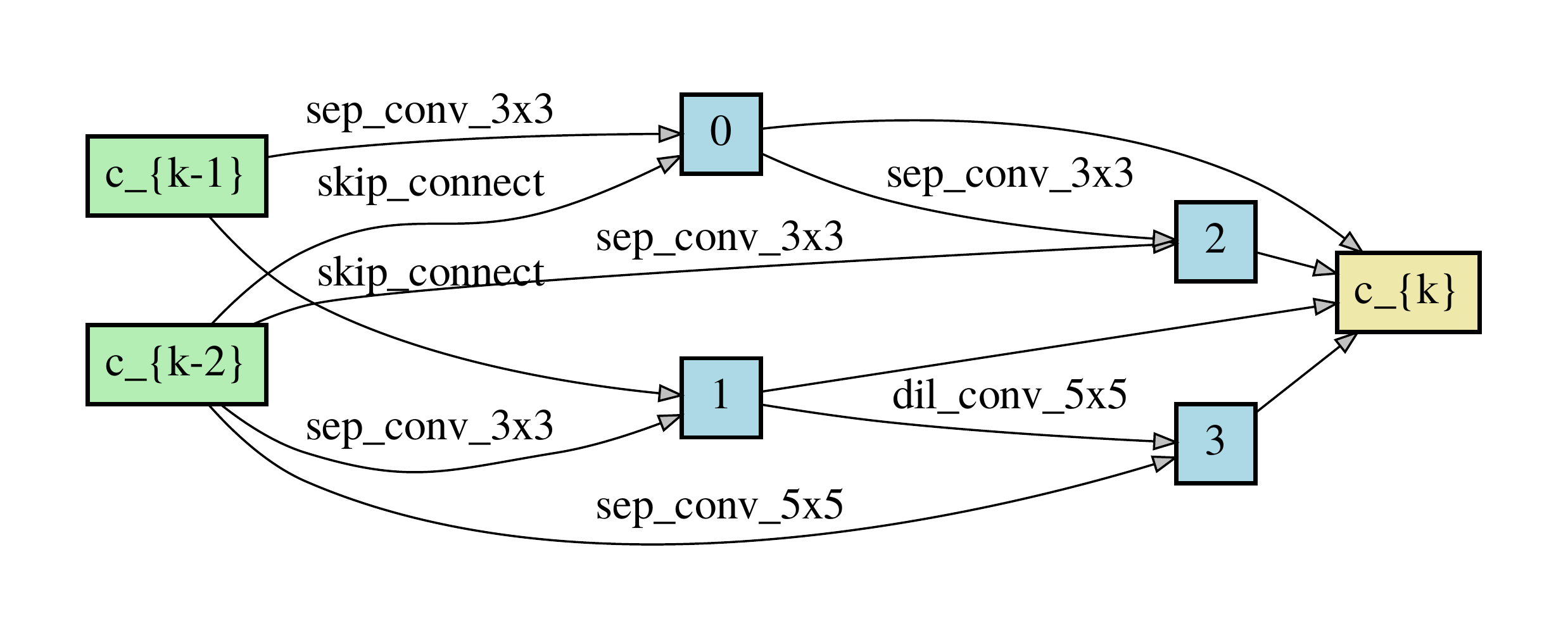}}
\hspace{0.2em}
\subfigure[The normal cell on ImageNet]{\label{pc_im_ncells_s3}\includegraphics[width=0.42\linewidth]{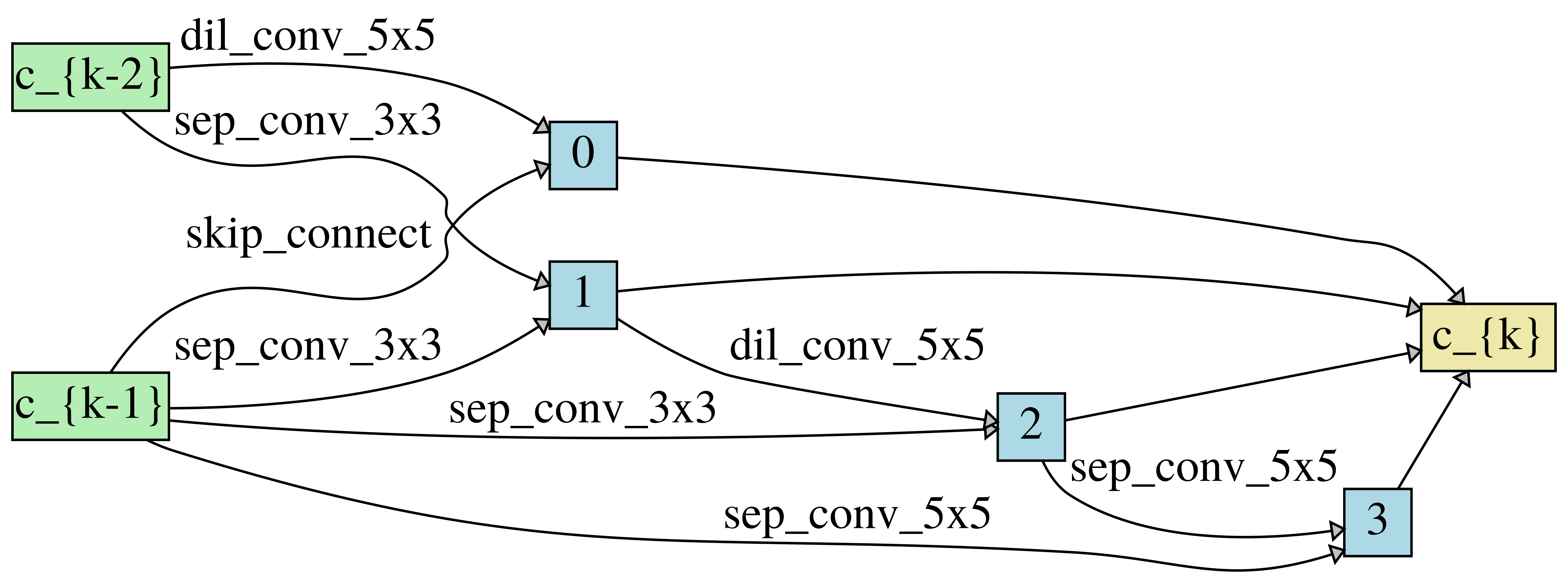}}
\subfigure[The reduction cell on CIFAR-10]{\label{pc_im_ncells_s2}\includegraphics[width=0.42\linewidth]{Fig/cifar_r.pdf}}
\hspace{0.2em}
\subfigure[The reduction cell on ImageNet]{\label{pc_im_ncells_s4}\includegraphics[width=0.42\linewidth]{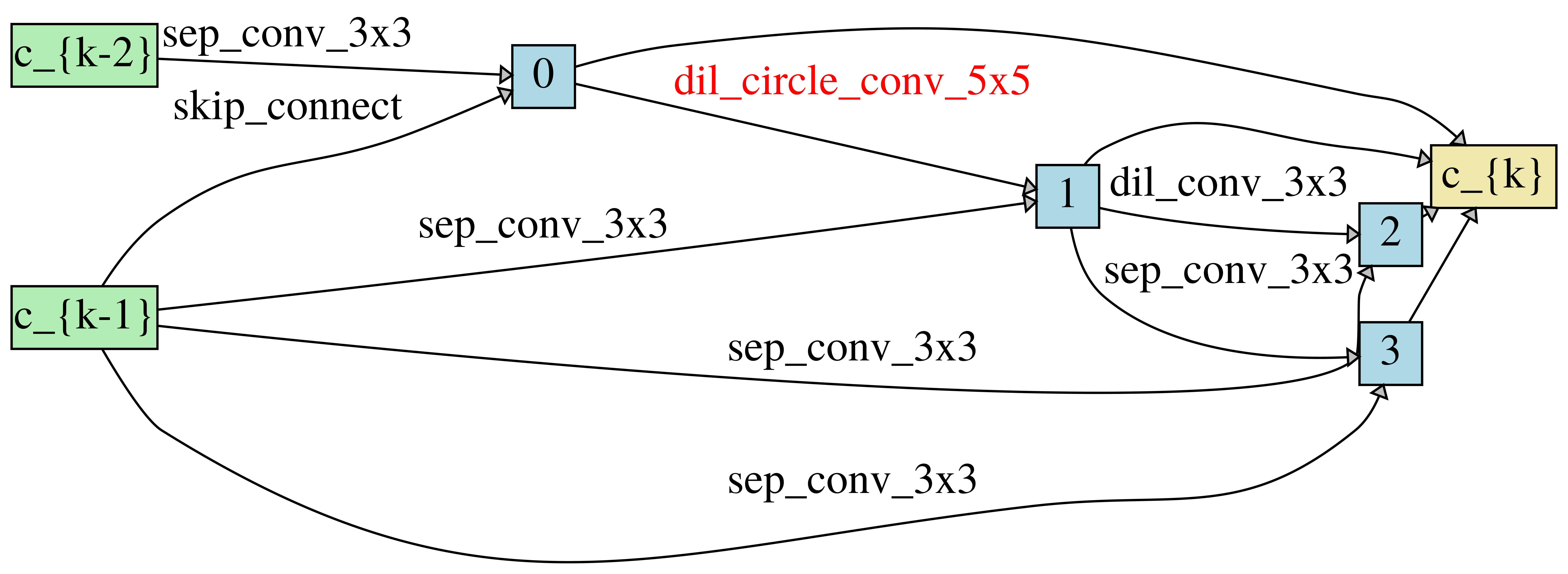}}
\caption{The searched normal cells and reduction cells of PC-DARTS-Circle on CIFAR-10 (\textit{left}) and ImageNet (\textit{right}). The circular kernels are marked in \red{red}}
\vspace{-0.5em}  
\label{fig:searched_cells_PC_DARTS}
\end{figure*}

%% file: 05Analysis.tex

For further analysis, we first show the better rotation invariance of large circular kernels compared to large square kernels by the evaluation on rotated or sheared images. Then, by constructing the integrated kernel, we reveal that the circular kernel has an optimization path different from the square kernel. In the Supplementary, 
we compare convolutions of large circular kernel with deformable convolutions~\citep{jeon2017active,dai2017deformable,zhu2019deformable} in NAS. The searched architecture containing convolutions of large circular kernels is with much less search and evaluation time cost and considerably outperforms the counterpart containing deformable convolutions.

\input{Figure/robustness}

\subsection{Rotation Invariance of Large Circular Kernels}
To better understand the approximate rotation-invariant property of circular kernels, we investigate their robustness to rotated images or sheared images.  
Specifically, we compare the performance of PC-DARTS-Circle with PC-DARTS searched on CIFAR-10. 
They are both trained on the training set of CIFAR-10 with standard data augmentation.
Fig.~\ref{fig:robustness} illustrates the classification errors on the rotated or sheared images generated on the test set of CIFAR-10. 
The rotation or shear angle range takes the value in $\sD_{e}=\{10, 20, 30, 40, 50, 60, 70, 80\}$. For each angle range $a \in \sD_{e}$, the images are rotated or sheared with an angle uniformly sampled from $\left( -a, a  \right)$, 
and we report the average classification error for three independent tests. \par

We can observe that the advantages of PC-DARTS-Circle steadily increase after $a > 30$, and reach the maximum at $a = 70$, which is roughly $4\%$ for rotation and $5\%$ for shear. The experiments not only justify the better rotation-invariant property of circular kernels, but also reveal that the rotation-invariant property of circular kernels in some layers is helpful to make the overall model more robust to the rotated images or sheared images.

\input{Figure/int_sc}

\subsection{Integrated Kernels and Optimal Path}
\label{sec:3x3-Kernel}

In Section~\ref{transformation}, we 
analyzed that the model with circular kernels has an optimal path different from that of the model with square kernels during the course of the gradient descent optimization. 
Here, we show their difference empirically by substituting all the $3 \times 3$ square kernels with the corresponding circular kernels or integrated kernels (explained in the followup paragraph) on WRNCifar~\citep{zagoruyko2016wide} and DenseNetCifar~\citep{huang2017densely}. 
\par

We first introduce the integrated kernel in detail. 
Each integrated kernel has two candidate kernels containing a square kernel and a circular kernel, denoted by $\sD = \{\sS,\sR\} $. At each iteration, we randomly select $\sD_p \in \sD $ according to a \textit{binomial} distribution for each convolutional layer. Following~\Eqref{eq.circle_conv} in Section~\ref{sec:method}, we resample the input $\mI$ with a group of offsets denoted by $\{\Delta\vd\}$ that corresponds to each discrete kernel position $\bm{s}$ to form the integrated receptive field.
Then, the output feature map of the corresponding convolution is defined as $\mO_{j} =  \sum_{\bm{s}\in \sS} \mW_{\bm{s}} \mI_{\vj+\bm{s}+\Delta\vd}$. 
The two types of kernels share the weight matrix but have distinct transformation matrices. During the training, the shared weight matrix is updated at each epoch, but the transformation matrices are randomly picked to determine the type of kernels of each layer at each iteration. \par

We compare the performance of the three versions of kernels on CIFAR-10 and CIFAR-100. 
The results are presented in Fig.~\ref{fig:int_sc}. 
It is worth noting that the version with integrated kernels yields better results over the other two versions, even though the network architecture is manually designed based on square kernels.
The superiority of the integrated kernels indicates that the switch between circular kernels and square kernels helps the model jump out of the local optima to perform better performance. Consequently, we can conclude 
that the circular kernel has an optimization path different from the square kernel for the gradient descent optimization.

%% file: Figure/robustness.tex

\begin{figure}[ht] 
\centering


\subfigure[\small{Rotation}]
{\label{rotate}\includegraphics[width=0.4\linewidth]{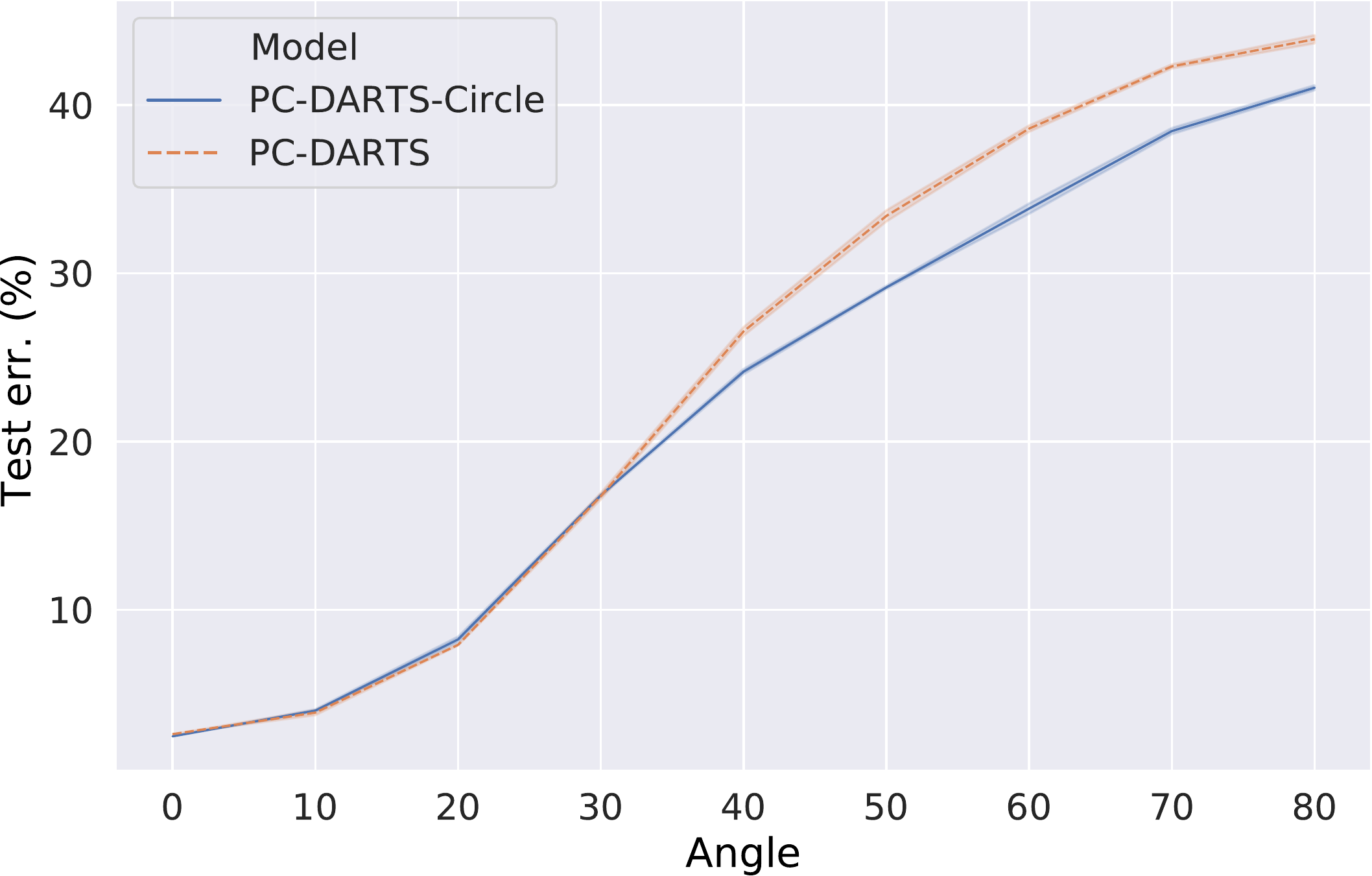}}
\hspace{10pt}
\subfigure[\small{Shear}]
{\label{shear}\includegraphics[width=0.4\linewidth]{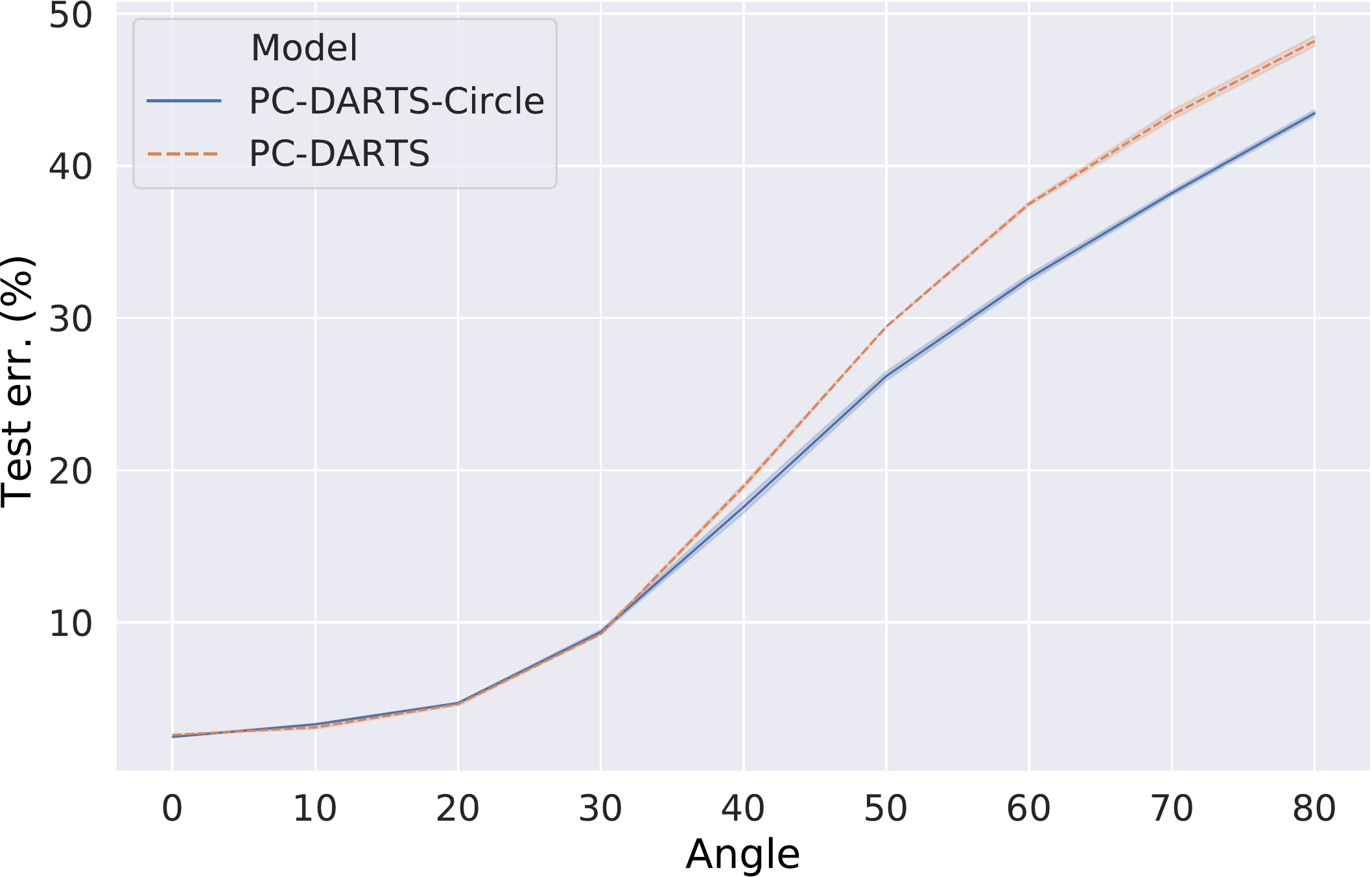}}
\caption{Comparison of classification error on rotated images 
(\textit{left}) or sheared images (\textit{right})}
\label{fig:robustness}
\end{figure}

%% file: Figure/int_sc.tex
\begin{figure*}[t]
\centering
\subfigure[\small{CIFAR-10}]
{\label{fig:ci10}\includegraphics[width=0.41\linewidth]{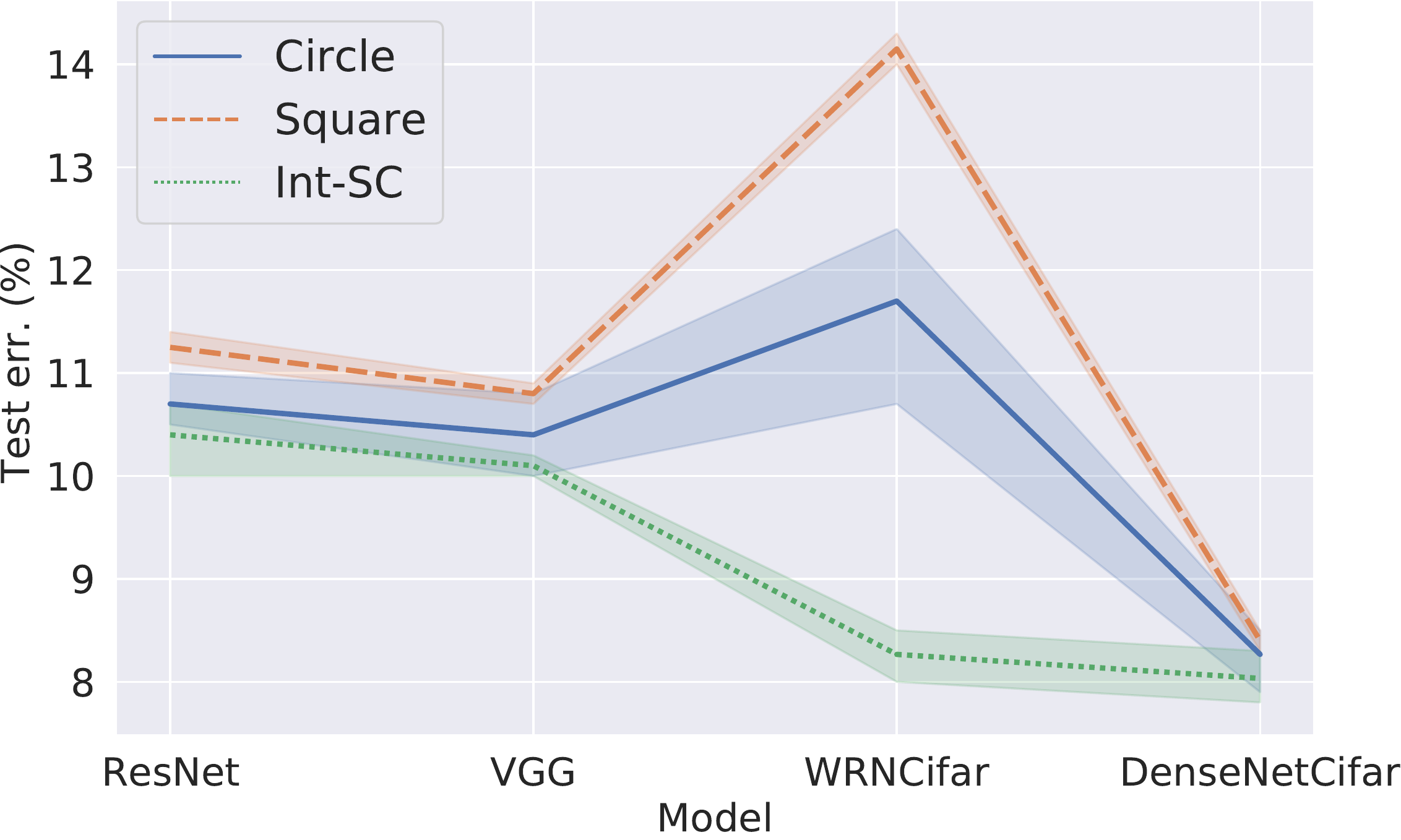}}
\hspace{10pt}
\subfigure[\small{CIFAR-100}]
{\label{fig:ci100}\includegraphics[width=0.41\linewidth]{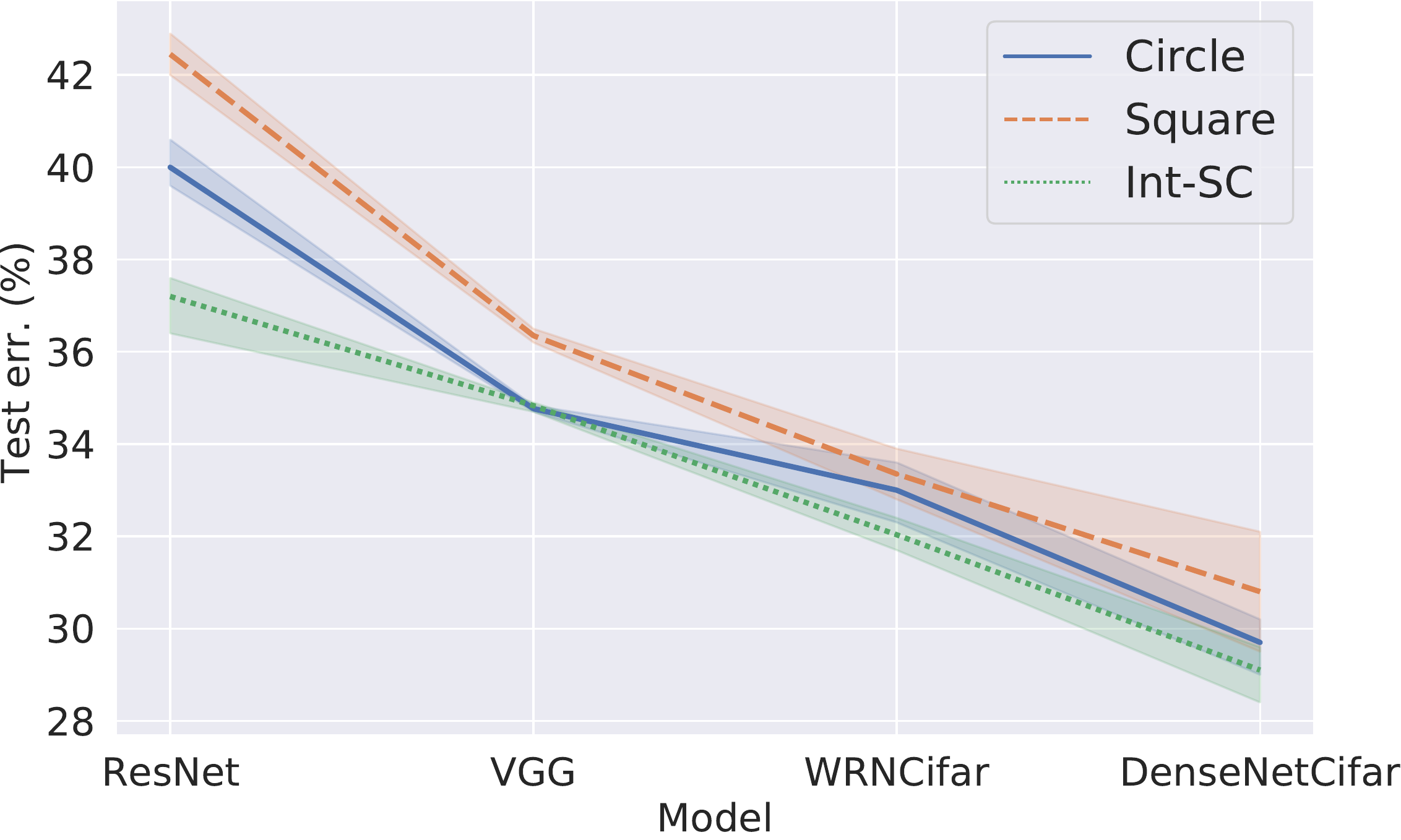}}
\caption{Test error ($\%$) of the baselines with square kernels (\textit{Square}) and the corresponding modified versions with circular kernels (\textit{Circle}) or integrated kernels (\textit{Int-SC}) on CIFAR-10 and CIFAR-100 without data augmentation (best viewed in color)}
\label{fig:int_sc}
\end{figure*}

%% file: 06Appendix.tex
The first part of this supplementary is a detailed analysis of the transformation matrix in relation to Section 3.3 of the main text. Then, using CIFAR-10 and CIFAR-100 with standard data augmentation, we show the advantages of large circular kernels over large square kernels, related to Section 4.1 of the main text. The benefits of big circular convolutions over deformable convolutions are next demonstrated, which is relevant to Section 5 of the main text. Finally, there are details on experiments and visualization.

\subsection{Analysis on the Transformation Matrix}
\label{appendix:transformation}
This section provides a theoretical analysis of the actual effect of the transformation matrix. 
Based on Equation $7$ in the main text,  for a circular convolution, the output feature map $\mO=\mW \otimes \left(\mB\star\mI\right)=\left(\mW \star \mB \right)\otimes\mI$. Then the squared value of a change on the output $\Delta \mO = \mO^{t+1}-\mO^t$ can be calculated as: 
\begin{equation}
\begin{aligned}
\| \Delta \mO \|^2 &=\left(\Delta \mW \otimes \left( \mB\star \mI \right)\right)^{\top}\left(\Delta \mW \otimes \left(\mB\star \mI \right)\right) \\
&=\left( \mB\star \mI \right)^{\top} \otimes \Delta \mW^{\top} \Delta \mW \otimes \left( \mB\star \mI \right),
\label{eq.o1}
\end{aligned}
\end{equation}
where $\Delta \mW$ is defined as $\mW^{t+1}-\mW^t$. Here the magnitude of $\Delta \mO$ is determined by the interaction between $\Delta\mW^{\top} \Delta \mW $ and $\mB \star \mI$, while $\Delta \widetilde\mO$ of the traditional convolutional layers is determined by $\Delta \mW^{\top} \Delta \mW$ and $\mI$. So the transformation matrix $\mB$ actually warps the receptive field in the input feature map $\mI$. And~\Eqref{eq.o1} can be transferred to: 
\begin{equation}
\begin{aligned}
\| \Delta \mO \|^2 &= \left(\left(\Delta \mW\star \mB\right) \otimes \mI \right)^{\top} \left( \left(\Delta \mW\star \mB \right) \otimes \mI \right) \\
&=   \mI^{\top} \otimes \left( \mB^{\top}\star \Delta\mW^{\top} \Delta \mW \star \mB  \right) \otimes\mI. 
\label{eq.o2}
\end{aligned}
\end{equation}
Here the magnitude of $\Delta \mO$ is determined by $\mB^{\top}\star \Delta\mW^{\top} \Delta \mW\star \mB$ and $\mI$, while $\Delta \widetilde\mO$ of traditional convolutional layers is determined by $\Delta \mW^{\top} \Delta \mW$ and $\mI$. So the transformation matrix $\mB$ can also be regarded as warping the kernel space. From ~\Eqref{eq.o1} and ~\Eqref{eq.o2}, we can conclude that the transformation matrix $\mB$ affects the optimal paths of gradient descent in both receptive field and kernel space.

\subsection{The Advantages of Large Circular Kernels}

We have shown that a large circular kernel has a more round receptive field and is more distinguishable from the corresponding square kernel. We conjecture that the larger circular kernels would exhibit more significant advantage over the square kernels if the circular kernels are helpful for deep learning tasks. To verify this hypothesis, we augment VGG~\citep{simonyan2014very}, ResNet~\citep{he2016deep}, WRNCifar~\citep{zagoruyko2016wide}, DenseNetCifar~\citep{huang2017densely}, and their circular kernel versions with larger kernel sizes and compare their performance on CIFAR-10 and CIFAR-100. For a fair comparison, we show the results of the original data without data augmentation in the main text. Here, we show the results on CIFAR-10 and CIFAR-100 with standard data augmentation. 

As shown in Table~\ref{large-circular-kernel-aug}, the performance of both the baselines and the corresponding circular kernel versions basically declines with the increment of kernel size, because the original neural network architecture is designed and hence optimized on the $3 \times 3$ square kernel. Nevertheless, we see that the advantage of circular kernels over square kernels becomes more distinct for larger kernels, indicating the superiority of large circular kernels. \par 
\input{Table/large_circular_cifar}

\subsection{Comparison with Deformable Convolution}
\label{sec:defor}
A natural idea of improving the operation space of NAS is to expand the search space with deformable convolutions~\citep{jeon2017active,dai2017deformable,zhu2019deformable}, as deformable convolutions are flexible and work well in manual architectures but are never considered in existing NAS methods. However, because NAS is a complicated bilevel optimization problem even on the standard convolutions~\citep{DBLP:conf/iclr/LiuSY19}, convolutional adaptation methods that require additional parameters are unlikely to be applicable to NAS due to the excessive optimization and considerable computation overhead. We show the performance of searched architecture containing deformable convolutions on CIFAR-10 in \ref{NAS-search-defor}. Compared to PC-DARTS-Circle, PC-DARTS-Deformable takes $7$ times search cost and $3$ times evaluation cost. Due to the time constraints, we only train PC-DARTS-Deformable for $100$ epochs, as the complete training for $600$ epochs requires $28.8$ GPU-days, which are $18$ times evaluation cost of PC-DARTS-Circle.

\input{Table/offset_vs_circle}

\subsection{More Details on Experiments}
\label{sec:experimental-details}

\subsubsection{Manually Designed Models on CIFAR Datasets}
\label{sec:CIFAR-10-Artificially}
This subsection provides additional details for comparing circular kernels versus square kernels by augmenting VGG~\citep{simonyan2014very}, ResNet~\citep{he2016deep}, WRNCifar~\citep{zagoruyko2016wide}, DenseNetCifar~\citep{huang2017densely}, with larger kernel sizes on CIFAR-10 and CIFAR-100 datasets. The two CIFAR datasets~\citep{krizhevsky2009learning} consist of colored natural images in 32$\times$32 pixels. The training set and test set contain $50,000$ and $10,000$ images, respectively. 
We train the models for $200$ epochs with batch size $128$ using weight decay $5 \times 10^{-4}$ and report the test error of the final epoch. No augmentation or standard data augmentation~\citep{he2016deep,huang2017densely,larsson2016fractalnet,lee2015deeply} (padding to 40$\times$40, random cropping, left-right flipping) is employed.
And we utilize the Stochastic Gradient Descent (SGD) optimizer with the momentum of 0.9. The learning rate initiates from 0.1 and gradually decays to zero following a half-cosine-function-shaped schedule with a warm-up at the first five epochs. 

\subsubsection{Searched Models on CIFAR-10}
\label{sec:CIFAR-10-NAS}

For the search and evaluation of DARTS-Circle and PC-DARTS-Circle on CIFAR-10, we follow the setup in DARTS and PC-DARTS. \par  
In the search scenario, we train the network for $50$ epochs. The $50\mathrm{K}$ training set of CIFAR-10 is split into two equal-sized subsets, with one subset used for training the network weights and the other used to search the architecture hyper-parameters. 
The network weights are optimized by momentum SGD, with a learning rate annealed down to zero following a cosine schedule without restart, a momentum of 0.9, and a weight decay of $3\times10^{-4}$. For the architecture hyper-parameters, we employ an Adam optimizer~\citep{kingma2014adam}, with a fixed learning rate of $6\times10^{-4}$, a momentum of $(0.5,0.999)$, a weight decay of $10^{-3}$, and initial number of channels $16$. 
Following DARTS, DARTS-Circle has a batch size of $64$ with an initial learning rate of $0.025$.
Following PC-DARTS, PC-DARTS-Circle has a batch size of $256$ with an initial learning rate of $0.1$.
In DARTS-Circle, we found that almost all edges in the derived normal cell are connected with $5 \times 5$ circular separable convolutions under the configuration of DARTS. Considering the strategy of the edge selection in DARTS is not very reasonable as reported by~\citep{chu2020fair}, we use the \textit{edge normalization} introduced in PC-DARTS to produce a more reasonable strategy of the edge selection.

In the evaluation stage, the models are trained from scratch for $600$ epochs with a batch size of $96$ and initial number of channels $36$. We apply the SGD optimizer with an initial learning rate of $0.025$ (annealed down to zero following a cosine schedule without restart), a momentum of $0.9$, a weight decay of $3\times10^{-4}$, and a norm gradient clipping at $5$. Cutout~\citep{devries2017improved}, as well as drop-path with a rate of $0.3$ are also used for regularization.

\subsubsection{Searched Models on ImageNet}
\label{sec:ImageNet-NAS}
The ILSVRC 2012 classification dataset~\citep{deng2009imagenet}, ImageNet, consists of $1.3M$ training images and $50K$ validation images, all of which are high-resolution and
roughly equally distributed over all the 1000 classes. The search and evaluation of DARTS-Circle and PC-DARTS-Circle on ImageNet follow DARTS~\citep{DBLP:conf/iclr/LiuSY19} and PC-DARTS. \par  
The search stage on ImageNet only exists in PC-DARTS-Circle. The model is trained for $50$ epochs with a batch size of 1024. The architecture hyper-parameters are frozen during the first $35$ epochs. For architecture hyper-parameters, we utilize the Adam optimizer~\citep{kingma2014adam} with a fixed learning rate of $6\times10^{-3}$, a momentum of $(0.5,0.999)$, and a weight decay of $10^{-3}$. For the network weights, we utilize a momentum SGD with the initial learning rate of $0.5$ (annealed down to zero following a cosine schedule without restart), a momentum of $0.9$, and a weight decay of $3\times10^{-5}$.

In the evaluation stage, the models are trained from scratch for $250$ epochs using a batch size of $512$ and the initial channel number $48$. We use the SGD optimizer with a momentum of $0.9$, an initial learning rate of $0.25$ (decayed down to zero linearly), and a weight decay of $3\times10^{-5}$. Additional enhancements are adopted, including label smoothing and an auxiliary loss tower during the training. The learning rate warm-up is applied for the first $5$ epochs.

\subsubsection{Visualization of the Searched Cells}
\label{sec:searched-cells}
In this section, we visualize the searched normal cells and reduction cells for DARTS and DARTS-Circle on CIFAR-10,
for PC-DARTS and PC-DARTS-Circle on CIFAR-10, and for PC-DARTS and PC-DARTS-Circle on ImageNet, in Fig.~\ref{fig:cells_darts}, Fig.~\ref{fig:cells_cifar}, and Fig.~\ref{fig:cells_imagenet}, respectively.
From the visualizations, we can observe that the normal cells of DARTS-Circle and PC-DARTS-Circle contain more large convolutions than those of the original versions. Additionally, convolutions of large circular kernels mainly exist in the reduction cells. According to DARTS~\citep{DBLP:conf/iclr/LiuSY19}, cells located at the $1/3$ and $2/3$ of the total depth of the network are reduction cells, in which all the operations adjacent to the input nodes are of stride two. We speculate that large circular kernels are significant when the size of feature maps change.

\input{Figure/searched_cells_darts}

\input{Figure/searched_cells_cifar}

\input{Figure/searched_cells_imagenet}

%% file: Table/large_circular_cifar.tex
\begin{table*}[tb]
\centering
\caption{Test error ($\%$) of the baselines with square kernels (Square) and the corresponding circular kernel (Circle) versions in kernel size $k \in \{3,5,7 \}$ on CIFAR-10 and CIFAR-100. With the increment of kernel size, the advantage of circular kernel over square kernel becomes more distinct}
\label{large-circular-kernel-aug}

\begin{threeparttable}[b]
\resizebox{\textwidth}{!}{
\begin{tabular}{lccrccr}
\toprule
\multirow{2}{*}{\textbf{Model}}&\multicolumn{3}{c}{\textbf{CIFAR-10}}&\multicolumn{3}{c}{\textbf{CIFAR-100}}\\
\cmidrule(lr){2-4}\cmidrule(lr){5-7} 
&Square&Circle&Test Err.$\downarrow$&Square&Circle&Test Err.$\downarrow$\\
\midrule
VGG ($3 \times 3$) &$  5.91
\pm0.04 $&$ 5.81\pm0.21 $&$\textbf{\BLUE{0.10}}$ 
&$ 25.19\pm0.12 $&$ 25.10\pm0.10 $&$\textbf{\BLUE{0.09}}$\\

VGG ($5 \times 5$) &$ 6.96\pm0.21 $&$ 6.72\pm0.11 $&$\textbf{\BLUE{0.24}}$ 
&$ 28.08\pm0.29 $&$ 28.02\pm0.06 $&$\textbf{\BLUE{0.06}}$\\

VGG ($7 \times 7$) &$  7.77\pm0.06 $&$ 7.62\pm0.09  $&$\textbf{\BLUE{0.15}}$ 
&$ 30.59\pm0.29 $&$ 29.99\pm0.32 $&$\textbf{\BLUE{0.60}}$\\ 
\midrule

ResNet ($3 \times 3$)&$  5.75\pm0.03 $&$ 5.74\pm0.06 $&$\textbf{\BLUE{0.01}}$ 
&$ 27.47\pm0.19 $&$ 27.56\pm0.09 $&$\textbf{\BLUE{-0.09}}$\\
ResNet ($5 \times 5$)&$  6.33\pm0.07 $&$ 6.13\pm0.16 $&$\textbf{\BLUE{0.20}}$ 
&$ 28.00\pm0.41 $&$ 27.87\pm0.26 $&$\textbf{\BLUE{0.13}}$\\
ResNet ($7 \times 7$)&$  6.82\pm0.35 $&$ 6.62\pm0.07 $&$\textbf{\BLUE{0.20}}$
&$ 29.11\pm0.25 $&$ 28.27\pm0.03 $&$\textbf{\BLUE{0.84}}$\\

\midrule  

WRNCifar ($3 \times 3$)     &$  4.21\pm0.06 $&$ 4.25\pm0.09  $&$\textbf{\BLUE{-0.04}}$ &$ 20.59\pm0.18 $&$ 21.10\pm0.32 $&$\textbf{\BLUE{-0.51}}$\\
WRNCifar ($5 \times 5$)     &$  4.58\pm0.14 $&$ 4.39\pm0.18  $&$\textbf{\BLUE{0.19}}$ &$ 21.24\pm0.14 $&$ 21.16\pm0.12 $&$\textbf{\BLUE{0.08}}$\\
WRNCifar ($7 \times 7$)     &$  5.18\pm0.16 $&$ 4.87\pm0.06  $&$\textbf{\BLUE{0.31}}$ &$ 22.44\pm0.15 $&$ 21.91\pm0.18 $&$\textbf{\BLUE{0.53}}$\\ 
\midrule
DenseNetCifar ($3 \times 3$)&$  5.05\pm0.11 $&$ 5.20\pm0.09 $&$\textbf{\BLUE{-0.15}}$ &$ 22.76\pm0.20 $&$ 22.62\pm0.17 $&$\textbf{\BLUE{0.14}}$\\
DenseNetCifar ($5 \times 5$)&$  5.19\pm0.12 $&$ 5.15\pm0.11 $&$\textbf{\BLUE{0.04}}$ &$ 23.31\pm0.41 $&$ 22.96\pm0.16 $&$\textbf{\BLUE{0.35}}$\\
DenseNetCifar ($7 \times 7$)&$  5.47\pm0.34 $&$ 5.36\pm0.03 $&$\textbf{\BLUE{0.11}}$&$ 23.64\pm0.19 $&$ 23.26\pm0.10 $&$\textbf{\BLUE{0.38}}$\\
\bottomrule
\end{tabular}
}
\end{threeparttable}

\end{table*}

%% file: Table/offset_vs_circle.tex
\begin{table*}[ht]
\centering
\caption{Comparison with searched network architecture containing deformable convolutions (PC-DARTS-Deformable) on CIFAR-10}
\begin{threeparttable}[b]
\resizebox{0.8\textwidth}{!}{
\begin{tabular}{@{}lcccccc@{}}
\toprule
\multirow{2}{*}{\textbf{Architecture}} & \textbf{Test Err.} & \textbf{Params} & \textbf{Search Cost} & \textbf{Evaluation Cost} \\
&                            \textbf{(\%)} & \textbf{(M)} & \textbf{(GPU-days)} &\textbf{(GPU-days)}\\
\midrule
PC-DARTS-Deformable + cutout & 4.28 & 5.7  & 0.7& 4.8 \\
PC-DARTS-Circle + cutout & 2.54 & 3.5  & 0.1& 1.4 \\
\bottomrule
\end{tabular}
}
\end{threeparttable}
\vspace{-5pt}

\label{NAS-search-defor}
\end{table*}

%% file: Figure/searched_cells_darts.tex
\begin{figure*}[!htb]
\centering
\subfigure[The normal cell of DARTS]{\includegraphics[width=0.45\linewidth]{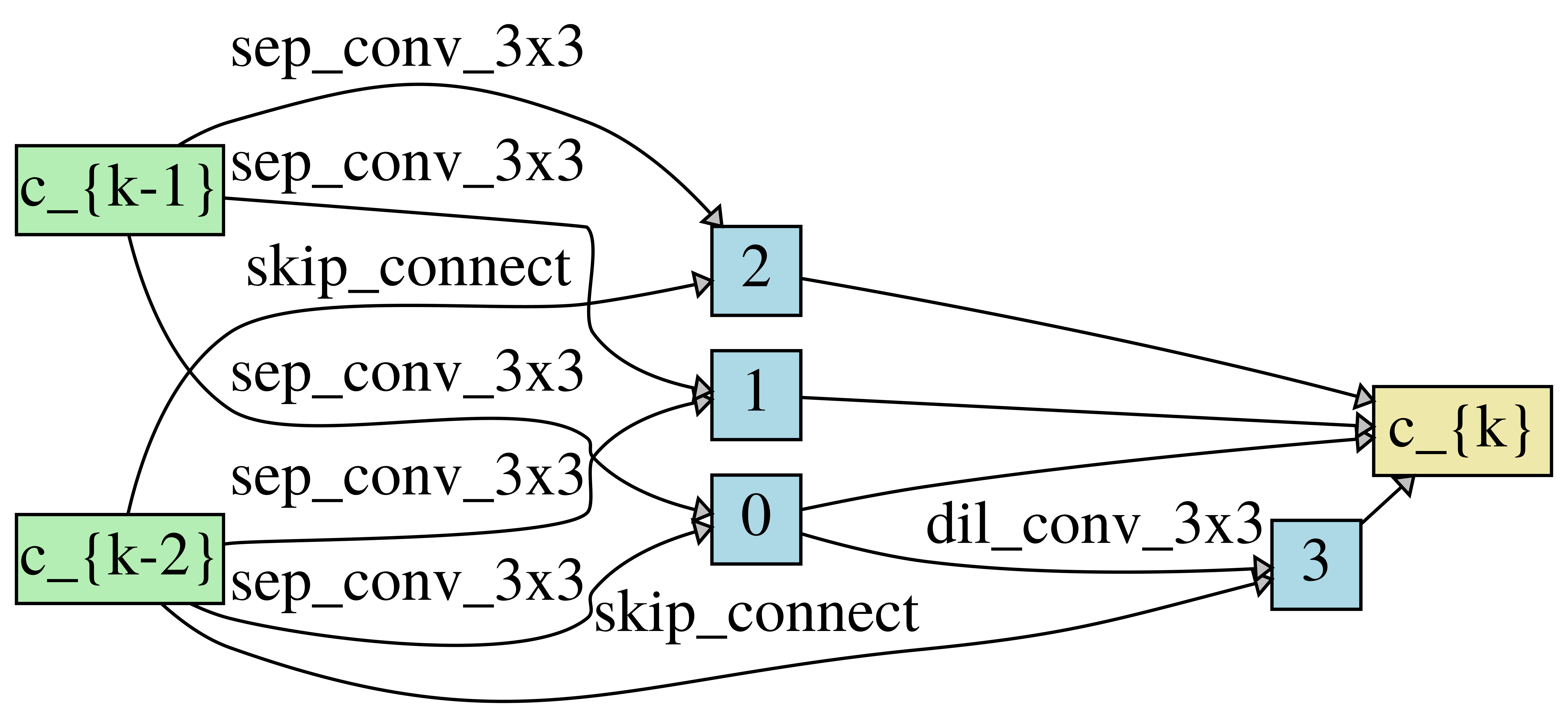}}
\hspace{0.2em}
\subfigure[The normal cell of DARTS-Circle]{\includegraphics[width=0.45\linewidth]{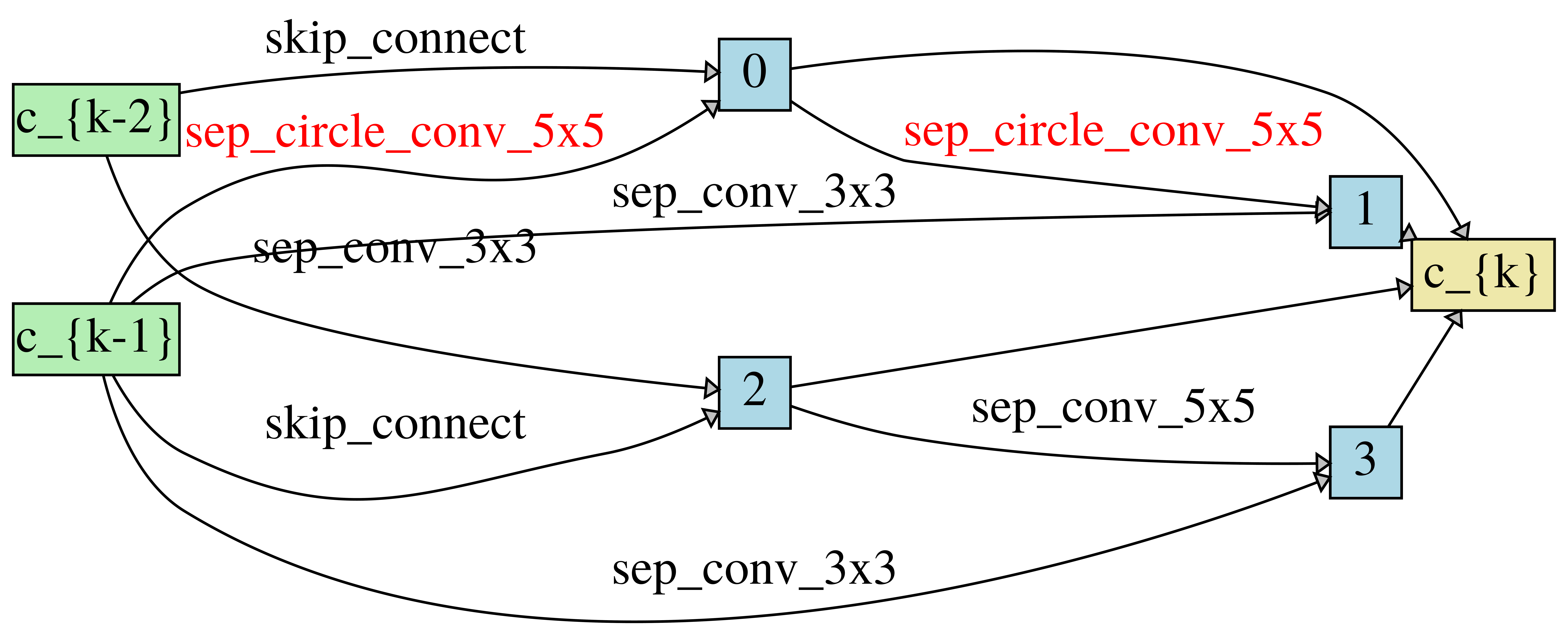}}
\subfigure[The reduction cell of DARTS]{\includegraphics[width=0.45\linewidth]{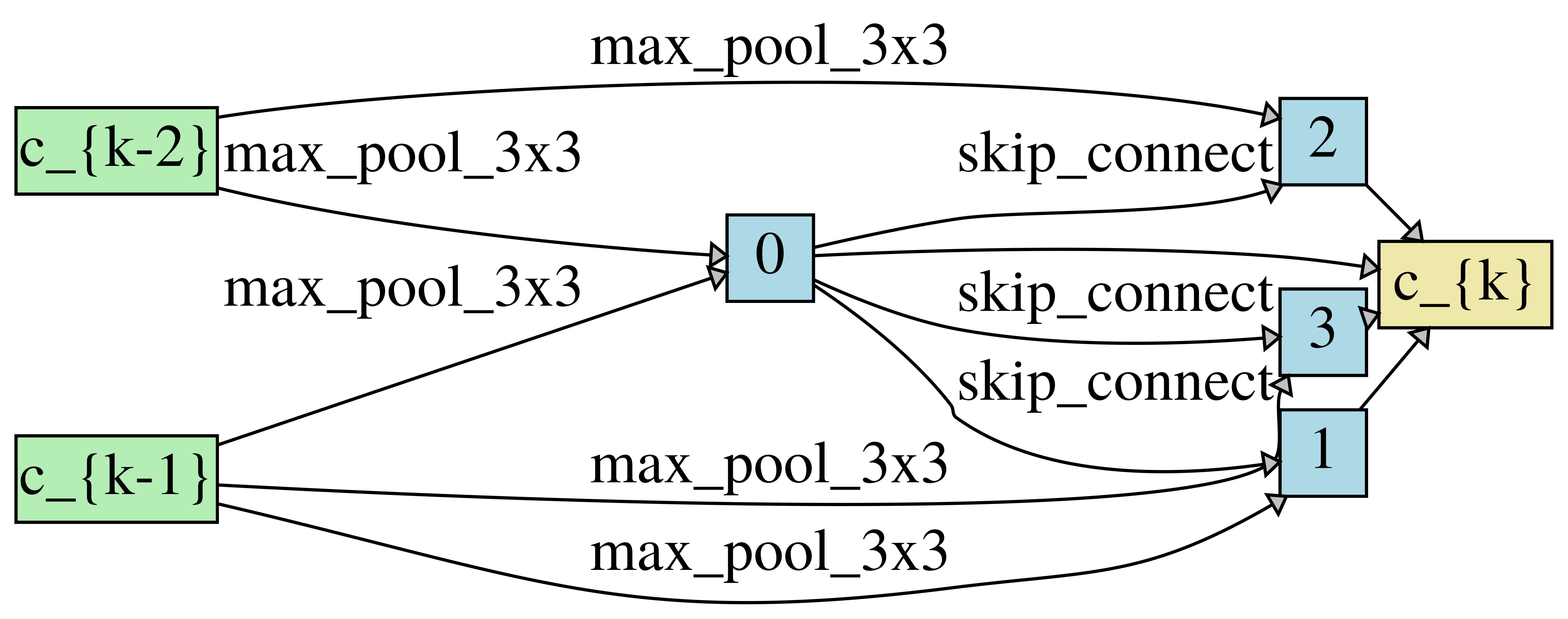}}
\hspace{0.2em}
\subfigure[The reduction cell of DARTS-Circle]{\includegraphics[width=0.45\linewidth]{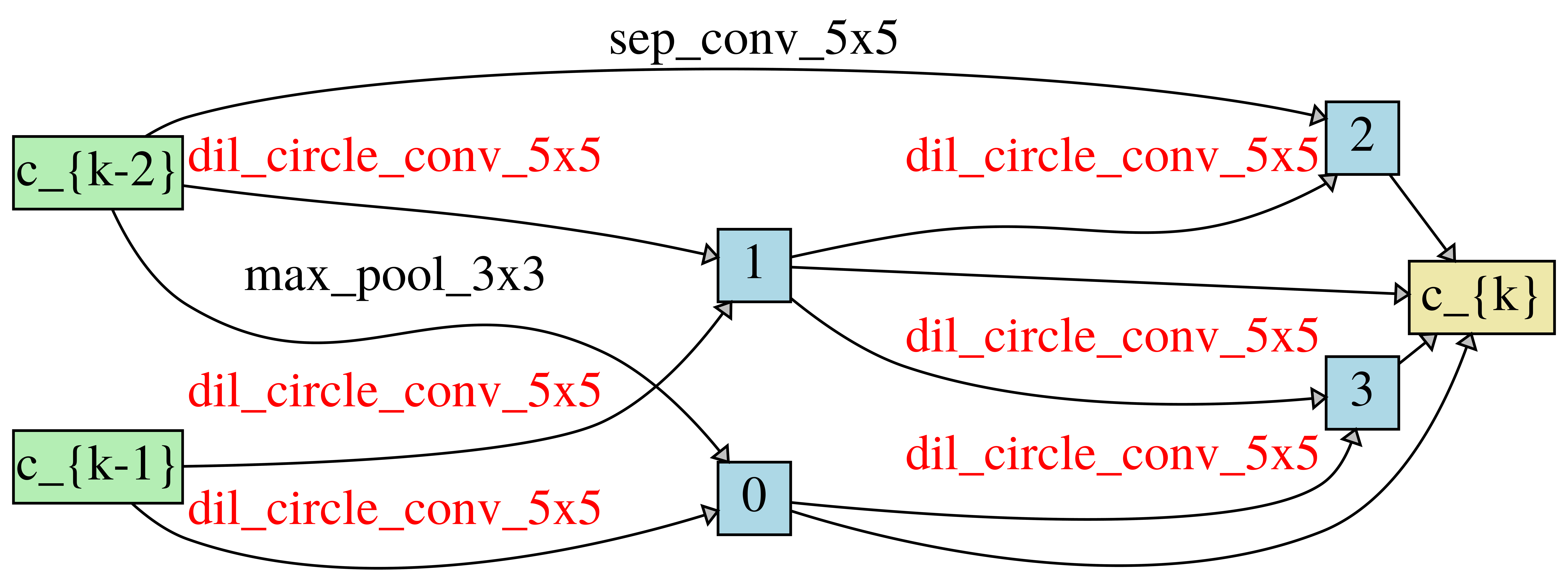}}
\caption{The searched normal and reduction cells of DARTS (\textit{left}) and DARTS-Circle (\textit{right}) on CIFAR-10. The large circular kernels are marked in \red{red}}
\label{fig:cells_darts}
\end{figure*}


%% file: Figure/searched_cells_cifar.tex
\begin{figure*}[!htb]
\centering
\subfigure[The normal cell of PC-DARTS]{\includegraphics[width=0.45\linewidth]{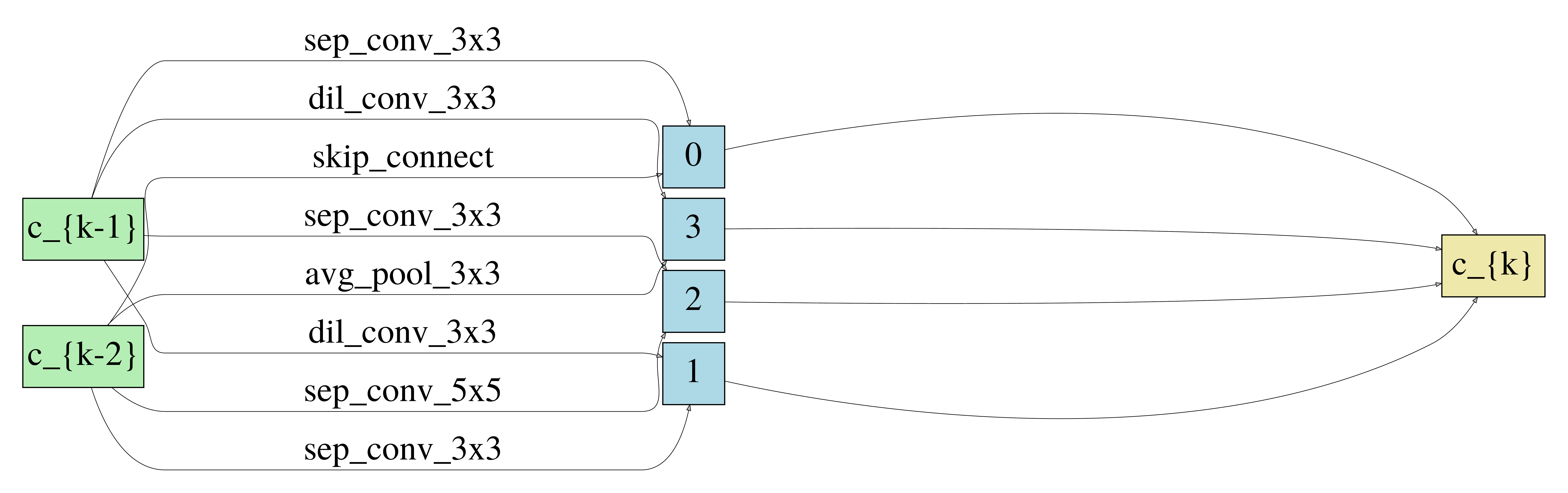}}
\hspace{0.2em}
\subfigure[The normal cell of PC-DARTS-Circle]{\includegraphics[width=0.45\linewidth]{Fig/cifar_n.pdf}}
\subfigure[The reduction cell of PC-DARTS]{\includegraphics[width=0.45\linewidth]{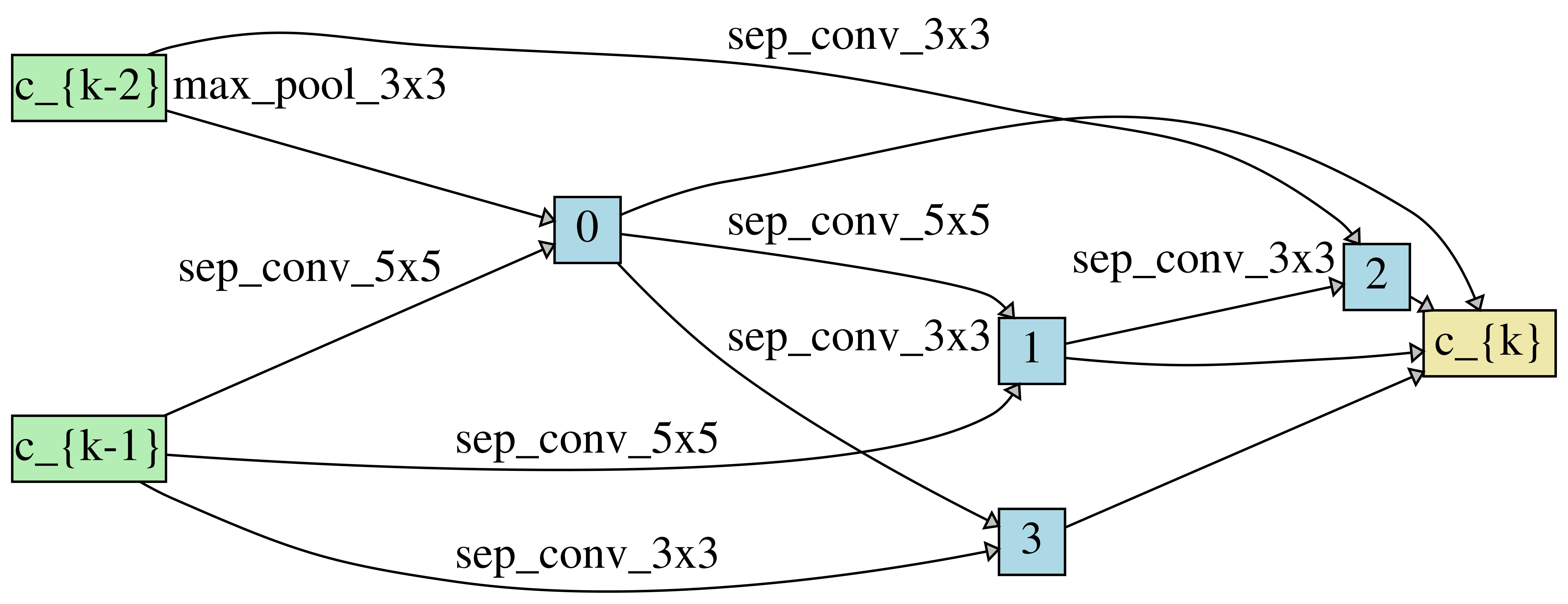}}
\hspace{0.2em}
\subfigure[The reduction cell of PC-DARTS-Circle]{\includegraphics[width=0.45\linewidth]{Fig/cifar_r.pdf}}

\caption{The searched normal and reduction cells of PC-DARTS (\textit{left}) and PC-DARTS-Circle (\textit{right}) on CIFAR-10. The large circular kernel is marked in \red{red}}
\label{fig:cells_cifar}
\end{figure*}



%% file: Figure/searched_cells_imagenet.tex
\begin{figure*}[!htb]
\centering
\subfigure[The normal cell of PC-DARTS]{\includegraphics[width=0.45\linewidth]{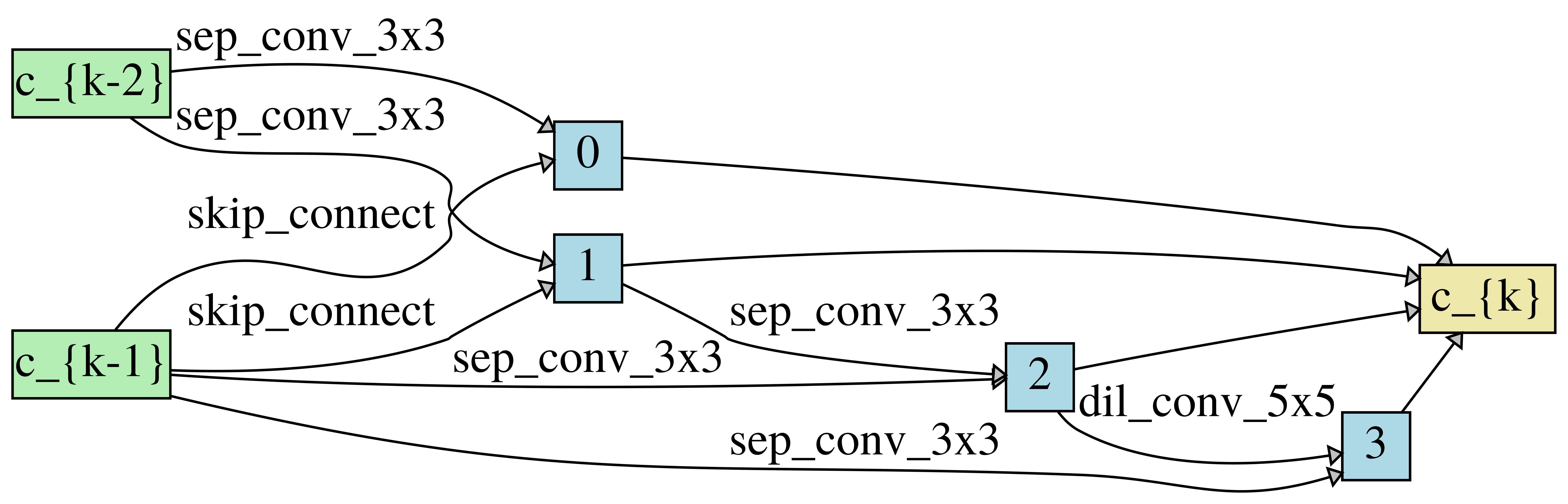}}
\hspace{0.2em}
\subfigure[The normal cell of PC-DARTS-Circle]{\includegraphics[width=0.45\linewidth]{Fig/imagenet_n.pdf}}
\subfigure[The reduction cell of PC-DARTS]{\includegraphics[width=0.45\linewidth]{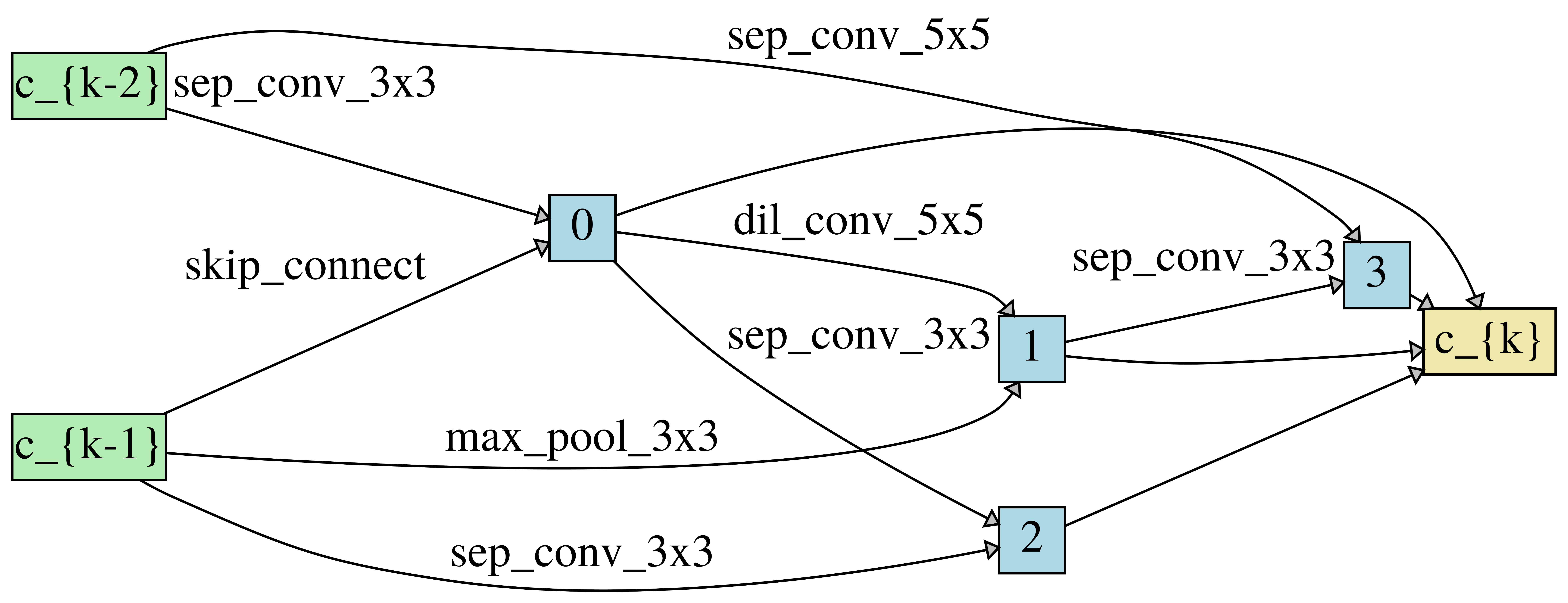}}
\hspace{0.2em}
\subfigure[The reduction cell of PC-DARTS-Circle]{\includegraphics[width=0.45\linewidth]{Fig/imagenet_r.pdf}}

\caption{The searched normal and reduction cells of PC-DARTS (\textit{left}) and PC-DARTS-Circle (\textit{right}) on ImageNet. The large circular kernels are marked in \red{red}}
\label{fig:cells_imagenet}
\end{figure*}
